\documentclass[10pt, dvipsnames]{article} % For LaTeX2e

\usepackage[accepted]{rlj} % Should be uncommented for the camera-ready
\usepackage{amssymb}            % Defines common symbols like \mathbb R
\usepackage{mathtools}          % Extends amsmath, providing common math tools
\usepackage{mathrsfs}           % Enables \mathscr, which can work in cases that \mathcal does not
\usepackage{graphicx}           % For including images
\usepackage{subcaption}         % Allows for the use of subfigures and subcaptions
\usepackage[space]{grffile}     % For spaces in image names
\usepackage{url}                % For displaying URLs
\usepackage{lipsum}             % For placeholder text

\usepackage{wrapfig}
\usepackage{subcaption}
\usepackage{algorithm}
\usepackage[noend]{algorithmic}
\usepackage{booktabs}
\usepackage{multirow}

\hypersetup{
    colorlinks = true,
    linkcolor = RubineRed,
    anchorcolor = OliveGreen,
    citecolor = OliveGreen,
    filecolor = OliveGreen,
    urlcolor = Plum
}

\DeclareMathOperator*{\argmin}{arg\,min}
\title{Eau De $Q$-Network: Adaptive Distillation of Neural Networks in Deep Reinforcement Learning
\vspace{-0.1cm}
}

\setrunningtitle{Eau De $Q$-Network}

\author{Théo Vincent\textsuperscript{1,2,$\dagger$} \quad Tim Faust\textsuperscript{1, 2} \quad Yogesh Tripathi\textsuperscript{1, 2} \\ Jan Peters\textsuperscript{1,2,3} \quad Carlo D'Eramo\textsuperscript{2,3,4}
}

\coverPageAuthor{Théo Vincent \quad Tim Faust \quad Yogesh Tripathi \\ Jan Peters \quad Carlo D'Eramo}

\emails{\vspace{-0.15cm}}

\affiliations{
$^{1}$DFKI GmbH, SAIROL $^{2}$Department of Computer Science, TU Darmstadt\\ $^{3}$Hessian.ai, TU Darmstadt $^{4}$Center for AI and Data Science, University of Wurzburg
\par % If including additional comments like below, use \par to add some whitespace. 
\vspace{-0.15cm}
$^\dagger$ correspondence to \url{theo.vincent@dfki.de}
\vspace{-0.5cm}
}

 % Your keywords

\summary{Recent works have successfully demonstrated that sparse deep reinforcement learning agents can be competitive against their dense counterparts. This opens up opportunities for reinforcement learning applications in fields where inference time and memory requirements are cost-sensitive or limited by hardware. To achieve a high sparsity level, the most effective methods use a dense-to-sparse mechanism where the agent's sparsity is gradually increased during training. Until now, those methods rely on hand-designed sparsity schedules that are not synchronized with the agent's learning pace. Crucially, the final sparsity level is chosen as a hyperparameter, which requires careful tuning as setting it too high might lead to poor performances. In this work, we address these shortcomings by crafting a dense-to-sparse algorithm that we name \emph{Eau De $Q$-Network}~(EauDeQN), where the online network is a pruned version of the target network, making the classical temporal-difference loss a distillation loss. To increase sparsity at the agent's learning pace, we consider multiple online networks with different sparsity levels, where each online network is trained from a shared target network. At each target update, the online network with the smallest loss is chosen as the next target network, while the other networks are replaced by a pruned version of the chosen network. Importantly, one online network is kept with the same sparsity level as the target network to slow down the distillation process if the other sparser online networks yield higher losses, thereby removing the need to set the final sparsity level. We evaluate the proposed approach on the Atari $2600$ benchmark and the MuJoCo physics simulator. Without explicit guidance, EauDeQN reaches high sparsity levels while keeping performances high. We also demonstrate that EauDeQN adapts the sparsity schedule to the neural network architecture and the training length. Our code is publicly available
at \url{https://github.com/theovincent/EauDeDQN} and the trained models are uploaded at \url{https://huggingface.co/TheoVincent/Atari_EauDeQN}.}

\contribution{We introduce \emph{Eau De $Q$-Network}~(EauDeQN), a dense-to-sparse reinforcement learning framework capable of adapting the sparsity schedule at the agent's learning pace while maintaining high performance. As a result, EauDeQN \textit{discovers} a final sparsity level. This means that EauDeQN avoids sparsity levels that are too high to yield high return and therefore removes the need to tune the final sparsity level.
}
{
Prior works in reinforcement learning consider hand-designed sparsity schedules and hard-coded final sparsity levels~\citep{graesser2022state}. EauDeQN is composed of Distill $Q$-Network (also introduced in this work, resembling \citet{ceronvalue}), which is responsible for gradually pruning the network during training, and Adaptive $Q$-Network~\citep{vincent2024adaqn}, which brings an adaptive behavior w.r.t. the agent's learning pace.
}

\begin{document}

\makeCover  % Create the cover page
\maketitle  % Make the title section

\begin{abstract}
Recent works have successfully demonstrated that sparse deep reinforcement learning agents can be competitive against their dense counterparts. This opens up opportunities for reinforcement learning applications in fields where inference time and memory requirements are cost-sensitive or limited by hardware. Until now, dense-to-sparse methods have relied on hand-designed sparsity schedules that are not synchronized with the agent's learning pace. Crucially, the final sparsity level is chosen as a hyperparameter, which requires careful tuning as setting it too high might lead to poor performances. In this work, we address these shortcomings by crafting a dense-to-sparse algorithm that we name \emph{Eau De $Q$-Network}~(EauDeQN). To increase sparsity at the agent's learning pace, we consider multiple online networks with different sparsity levels, where each online network is trained from a shared target network. At each target update, the online network with the smallest loss is chosen as the next target network, while the other networks are replaced by a pruned version of the chosen network. We evaluate the proposed approach on the Atari $2600$ benchmark and the MuJoCo physics simulator, showing that EauDeQN reaches high sparsity levels while keeping performances high.
% Comment for anonymous submission
\vspace{-0.35cm}
\end{abstract}

\section{Introduction} \label{S:introduction}
% Comment for anonymous submission
\vspace{-0.1cm}
Training large neural networks in reinforcement learning~(RL) has been demonstrated to be harder than in the fields of computer vision and natural language processing~\citep{henderson2018deep,ota2024training}. It is only in the last years that the RL community developed algorithms capable of training larger networks leading to performance increase~\citep{espeholt2018impala, schwarzer2023bigger, bhatt2019crossq, nauman2024bigger}. In reaction to those breakthroughs and inspired by the success of sparse neural networks in other fields~\citep{han2015deep, zhu2018prune, mocanu2018scalable, liu2020dynamic, evci2020rigging, frankesample}, recent works have attempted to apply pruning algorithms in RL to achieve well-performing agents composed of fewer parameters~\citep{yu2019playing, sokar2021dynamic, tanrlx2, ceronvalue}. Reducing the number of parameters promises to lower the cost of deploying RL agents. It is also essential for embedded systems where the agent's latency and memory footprint are a hard constraint.

Imposing sparsity in RL is not straightforward~\citep{liurethinking, graesser2022state}. The main objective is to reach high sparsity levels using as few environment interactions as possible. Therefore, we focus on methods that are gradually increasing the sparsity level, so-called \textit{dense-to-sparse} training, as they generally perform better than \textit{sparse-to-sparse} training methods, where the network is pruned before the training starts~\citep{arnob2021single, sokar2021dynamic, tanrlx2}. Most of those approaches borrow pruning techniques from the field of supervised learning, which were not designed for handling the specificities of RL~\citep{sokar2021dynamic, tanrlx2, ceronvalue}. Strikingly, those techniques impose pruning schedules that are not synchronized with the agent's learning pace. Additionally, the final sparsity level is a hyperparameter of the pruning algorithm, which is not convenient as its value is hard to predict and depends on the reinforcement learning setting, the task, the network architecture, and the training length~\citep{evci2019difficulty}.

\begin{figure}
    \centering
    \includegraphics[width=0.9\textwidth]{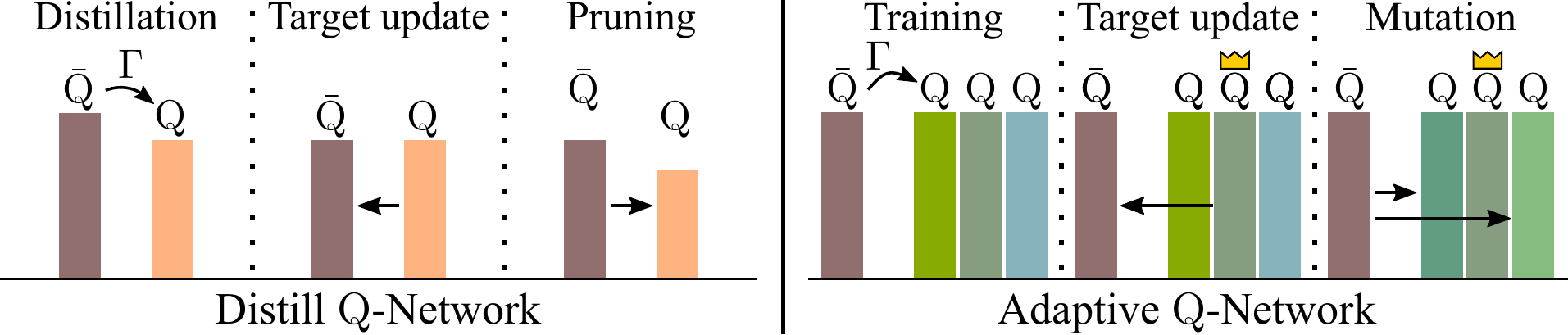}
    \caption{\textbf{Left}: In Distill $Q$-Network, the online network $Q$ is a pruned version of the target network $\bar{Q}$, transforming the classical temporal-difference loss, using the Bellman operator $\Gamma$, into a distillation loss. \textbf{Right}: Adaptive $Q$-Network~\citep{vincent2024adaqn} uses several online networks, each one defined with different hyperparameters and trained from a shared target network $\bar{Q}$. At each target update, the online network $Q$ with the lowest cumulated loss (represented with a crown) is chosen as the next target network. The target network is then copied to replace the other networks and the hyperparameters of the crowned network are mutated.}
    \label{F:distillqn_adaqn}
\end{figure}
\begin{wrapfigure}{r}{0.45\textwidth}
    \includegraphics[width=0.45\textwidth]{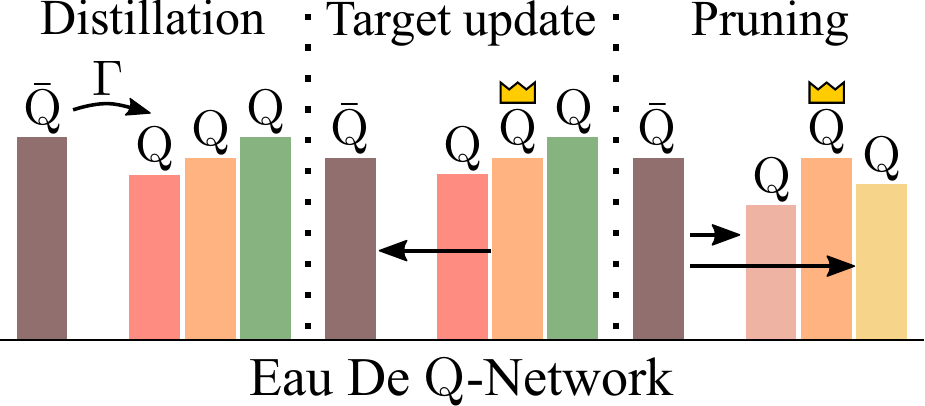}
    \caption{\emph{Eau De $Q$-Network} is based on Distill $Q$-Network (Figure~\ref{F:distillqn_adaqn}, left) and uses the adaptive ability of Adaptive $Q$-Network (Figure~\ref{F:distillqn_adaqn}, right) to prune the weights of the neural network at the agent's learning pace.}
    \label{F:idea}
\end{wrapfigure}
In this work, we propose a novel approach to sparse training that gradually prunes the weights of the neural networks at the agent's learning pace to finish at a sparsity level that is discovered by the algorithm. This behavior is made possible thanks to the combination of two independent methods, namely Distill $Q$-Network and Adaptive $Q$-Network~\citep{vincent2024adaqn}, gathered in a single algorithm coined \emph{Eau De $Q$-Network}~(EauDeQN). In Distill $Q$-Network~(DistillQN), the online network is a pruned version of the target network, thereby using the common temporal-difference loss as a distillation loss (Figure~\ref{F:distillqn_adaqn}, left). While DistillQN is also a novel approach to sparse training introduced in this work, it still relies on a hand-designed pruning schedule. This is why we will mainly focus on EauDeQN. Adaptive $Q$-Network~(AdaQN) was originally introduced to tune the RL agent's hyperparameters~(\citet{vincent2024adaqn}, Figure~\ref{F:distillqn_adaqn}, right). When combined with DistillQN, the resulting algorithm considers several online networks with equal or higher sparsity levels than the target network. At each target update, the online network with the lowest cumulated loss, represented with a crown in Figure~\ref{F:idea}, is chosen as the next target network. Therefore, the ability to select between different sparsity levels according to the value of the cumulated loss synchronizes the pruning schedule to the agent's learning pace. The target network is then copied and pruned to replace the other networks. Crucially, by keeping the online network with the lowest cumulated loss for the next iteration, the algorithm can keep the sparsity level steady if the cumulated loss increases for higher sparsity levels. This results in an algorithm capable of discovering the final sparsity level as opposed to the current methods, which require multiple training iterations to tune the final sparsity level.

\section{Background}
\textbf{Deep $Q$-Network~\citep{mnih2015human}} ~~ In a sequential decision-making problem, the optimal action-value function $Q^*(s, a)$ is the optimal expected sum of discounted future reward, given a state $s$, an action $a$. From this quantity, the optimal policy $\pi^*$ yielding the highest sum of discounted reward can be obtained by directly maximizing $Q^*(s, \cdot)$ for a given state $s$. Importantly, the optimal action-value function is the fixed point of the Bellman operator, which is a contraction mapping. The fixed point theorem guarantees that iterating endlessly over any $Q$-function with the Bellman operator converges to the fixed point $Q^*$. This is why, to compute $Q^*$, \citet{ernst05a} proposes to learn the successive Bellman iterations using an online network $Q$ and a target network $\bar{Q}$ representing the previous Bellman iteration. The training loss relies on the temporal-difference error, which is defined from a sample $(s, a, r, s')$ as
\begin{equation}
    \mathcal{L}_{\text{QN}}(Q) = (r + \gamma \max_{a'} \bar{Q}(s', a') - Q(s, a))^2,
\end{equation}
where QN stands for $Q$-Network. After a predefined number of gradient steps, the target network is updated to represent the next Bellman iteration. This procedure repeats until the training ends. \citet{mnih2015human} adapts this framework to the online setting where the agent interacts with the environment using an $\epsilon$-greedy policy~\citep{sutton1998rli} computed from the online network $Q$.

\textbf{Adaptive $Q$-Network~\citep{vincent2024adaqn}} ~~ The hyperparameters of DQN are numerous and hard to tune. This is why \citet{vincent2024adaqn} introduced AdaQN, which is designed to adaptively select DQN's hyperparameters during training. This is done by considering several online networks trained with different hyperparameters and sharing a single target network. At each target update, the online network with the lowest cumulated loss is selected as the next target network. After each target update, the selected online network is copied to replace the other online networks, and genetic mutations are applied to the hyperparameters of each copy to explore the space of hyperparameters.

\section{Related Work}
As discussed in Section~\ref{S:introduction}, we focus on dense-to-sparse training methods in this work, as they generally perform better than sparse-to-sparse methods~\citep{graesser2022state}. Sparse-to-sparse methods prune a dense network before the training starts~\citep{arnob2021single}, relying on the lottery ticket hypothesis~\citep{frankle2018lottery}. The lottery ticket hypothesis makes the assumption that, when initializing a dense network, there exist sub-networks that can lead to similar performances as the dense network if given similar resources. For such methods, the network morphology can still be adapted during training using gradient information~\citep{tanrlx2}, or evolutionary methods~\citep{sokar2021dynamic, grooten2023automatic}. On the other hand, dense-to-sparse methods start with a dense network and prune its connections during training. In the machine learning literature, we find approaches using variational dropout to sparsify the network~\citep{molchanov2017variational}. Alternatively, \citet{liurethinking} design learnable masks by approximating the gradient of the loss function w.r.t. the sparsity level with a piecewise polynomial estimate. Nonetheless, those approaches have not been adapted to an RL setting yet. In the RL literature, \citet{yu2019playing} evaluates the lottery ticket hypothesis in an RL setting using a hand-designed geometric sparsity schedule. They conclude that the lottery ticket hypothesis is only valid for a subset of Atari games~\citep{bellemare2013arcade}. Notably, Figure~$5$ in \citet{yu2019playing} shows that the performances greatly depend on the imposed final sparsity level. \citet{livne2020pops} makes use of a pre-trained teacher to boost performances. Distillation techniques using pre-trained networks have also been used to kickstart the training in single-task RL settings~\citep{zhang2019accelerating} and multi-task RL settings~\citep{schmitt2018kickstarting}.

\citet{ceronvalue} is the closest work to our approach. The authors gradually prune the neural network weights during training. They use a polynomial pruning schedule introduced in~\citet{zhu2018prune} and demonstrate that it is effective on the Atari~\citep{bellemare2013arcade} and MuJoCo~\citep{todorov2012} benchmarks, yielding even higher performances than the dense counterpart for \textit{wide} neural networks. As the authors do not give a name to their method, we will refer to it as Polynomial Pruning $Q$-Networks~(PolyPruneQN). At any time $t$ during training, a binary mask filtering out the weights with lowest magnitude imposes the sparsity $s_t$ to the neural network. $s_t$ is defined as 
\begin{equation} \label{E:polyprune}
    s_t = s_F \left(1 - \left(1 - \text{Clip}\left( \frac{t - t_{\text{start}}}{t_{\text{end}} - t_{\text{start}}}, 0, 1 \right) \right)^n \right),
\end{equation}
where $s_F$ corresponds to the final sparsity level, $t_{\text{start}}$ is the first timestep where the pruning starts, $t_{\text{end}}$ is the timestep after which the sparsity level is kept constant at $s_F$, and $n$ controls the steepness of the pruning schedule. One shortcoming of this approach is that those hyperparameters need to be tuned by hand for each RL setting, task, network architecture, and training length.

\section{Eau De Q-Network}
Our approach uses AdaQN's adaptivity to learn a pruning schedule synchronized with the agent's learning pace, therefore avoiding the need to impose a hard-coded sparsity schedule and final sparsity level. For that, we first introduce a novel algorithm called Distill $Q$-Network~(DistillQN), which resembles DQN except for the fact that after each target update, the online network is pruned as shown in Figure~\ref{F:distillqn_adaqn} (left). This algorithm belongs to the dense-to-sparse training family and relies on a hand-designed pruning schedule. We remark that PolyPruneQN is an instance of DistillQN when PolyPruneQN's pruning period is synchronized with the target update period. 

\begin{algorithm}[t]
\caption{Eau De Deep $Q$-Network (EauDeDQN). Modifications to DQN are marked in \textcolor{BlueViolet}{purple}.}
\label{A:eaudedqn}
\begin{algorithmic}[1]
\STATE Initialize \textcolor{BlueViolet}{$K$} online parameters $(\theta^k)_{k = 1}^K$, and an empty replay buffer $\mathcal{D}$. \textcolor{BlueViolet}{Set $\psi = 0$} and $\bar{\theta} \leftarrow \theta^{\psi}$ the target parameters. \textcolor{BlueViolet}{Set the cumulated losses $L_k = 0$, for $k=1, \ldots, K$.}
\STATE \textbf{repeat}
\begin{ALC@g}
\STATE \textcolor{BlueViolet}{Set $\psi^b \sim \texttt{Choice}(\{1, .., K\}, p=\{\frac{1}{L_1}, .., \frac{1}{L_K}\})$. Illustrated in Figure~\ref{F:exploration_exploitation_sampling} (right).}
\STATE Take action $a \sim \epsilon\text{-greedy} \label{L:action_sampling}(Q_{\theta^{\textcolor{BlueViolet}{\psi^b}}}(s, \cdot))$; Observe reward $r$, next state $s'$.
\STATE Update $\mathcal{D} \leftarrow \mathcal{D} \bigcup \{(s, a, r, s')\}$.
\STATE \textbf{every $G$ steps}
\begin{ALC@g}
\STATE Sample a mini-batch $\mathcal{B} = \{ (s, a, r, s') \}$ from $\mathcal{D}$.
\STATE Compute the \textit{shared} target $y \leftarrow r + \gamma \max_{a'} Q_{\bar{\theta}}(s', a')$.
\STATE \textbf{for} {\textcolor{BlueViolet}{$k=1, ..., K$}} \textbf{do} \textcolor{BlueViolet}{\textit{[in parallel]}}
\begin{ALC@g}
\STATE Compute the loss w.r.t $\theta^k$, $\mathcal{L}^k_{\text{QN}} = \sum_{(s, a, r, s') \in \mathcal{B}} \left( y - Q_{\theta^k}(s, a) \right)^2$.
\STATE Update $\theta^k$ from $\nabla_{\theta^k} \mathcal{L}^k_{\text{QN}}$ and \textcolor{BlueViolet}{$L_k \leftarrow L_k + \mathcal{L}^k_{\text{QN}}$}.
\end{ALC@g}
\end{ALC@g}
\STATE \textbf{every $T$ steps}
\begin{ALC@g}
\STATE Update the target network $\bar{\theta} \leftarrow \theta^{\psi}$, \textcolor{BlueViolet}{where $\psi \leftarrow \argmin_{k} L_k$}.
\STATE \textcolor{BlueViolet}{\texttt{Exploitation:} Select $K$ networks with repetition from the current population using the cumulated losses $L_k$. The process is illustrated in Figure~\ref{F:exploration_exploitation_sampling} (left).} \label{L:exploitation}
\STATE \textcolor{BlueViolet}{\texttt{Exploration:} Prune the duplicated networks at a sparsity level defined in Equation~\ref{E:eaudeqn}. The process is illustrated in Figure~\ref{F:exploration_exploitation_sampling} (middle).}
\STATE \textcolor{BlueViolet}{Reset $L_k \leftarrow 0$, for $k \in \{1, \ldots, K\}$.}
\end{ALC@g}
\end{ALC@g}
\end{algorithmic}
\end{algorithm}
\begin{figure}
    \includegraphics[width=\textwidth]{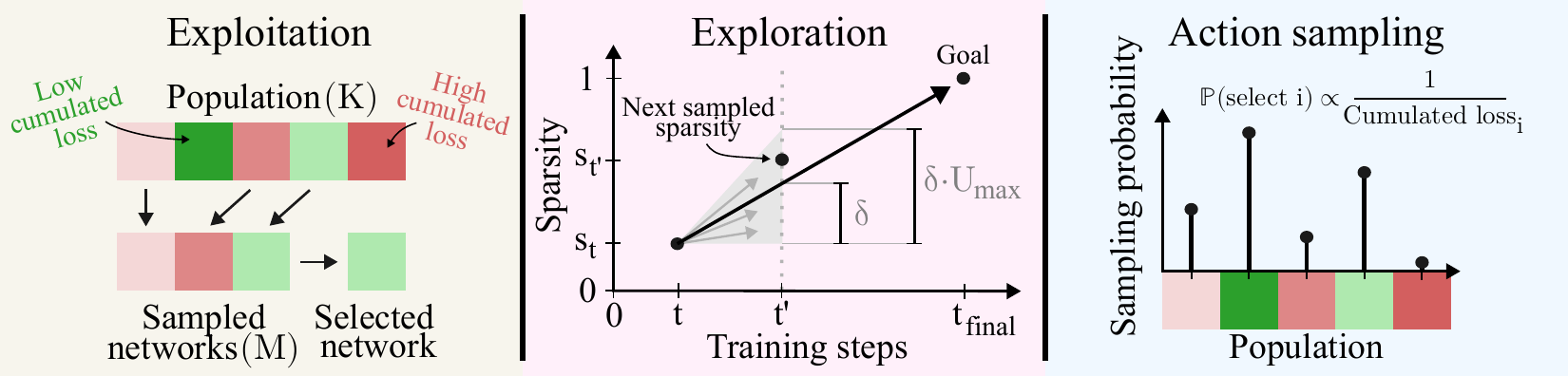}
    \caption{\textbf{Left:} The exploitation phase consists of selecting $K$ networks with repetition from the current population. Each network is selected as the one with the lowest cumulated loss out of $M$ uniformly sampled networks. \textbf{Middle:} In the exploration phase, new sparsity levels are sampled along the line joining the current point $(t, s_t)$ and the goal $(t_{\text{final}}, 1)$. To enhance exploration, the obtained sparsity is scaled by $U \sim \mathcal{U}(0, U_{\text{max}})$. \textbf{Right:} At each environment interaction, a network is selected from a probability distribution that is inversely proportional to the cumulated loss.}
    \label{F:exploration_exploitation_sampling}
\end{figure}
The combination of DistillQN and AdaQN, which we call Eau De $Q$-Network~(EauDeQN), adaptively selects the sparsity level based on the agent's learning pace. Indeed, EauDeQN considers $K$ online networks with different sparsity levels, each trained against a shared target network. Following the AdaQN algorithm, at each target update, the online network with the lowest cumulated loss is selected as the next target network (see Figure~\ref{F:idea}). Therefore, at each target update, the online network with the sparsity level that has been the most adapted to the optimization landscape related to the given loss function is selected as the next target network. Before the training continues, the cumulated loss of each online network is used to select the new population of $K$ online networks that will be used to continue the training. Inspired by \citet{miller1995genetic}, and \citet{frankesample}, each member of this new population is selected by randomly sampling $M$ online networks from the $K$ online networks and choosing the one with the lowest cumulated loss as illustrated in Figure~\ref{F:exploration_exploitation_sampling} (left). One spot in the new population is reserved for the online network chosen as the next target network, i.e., the one with minimal cumulated loss. This is referred to as the exploitation phase in Algorithm~\ref{A:eaudedqn}, Line~\ref{L:exploitation} as it filters out the online networks with a sparsity level that was not well suited for minimizing the current loss function. Then, an exploration phase is responsible for sampling a new sparsity level for each duplicated network. The new sparsity levels chosen at timestep $t$, are kept until timestep $t' = t + T$, which corresponds to the timestep of the following target update. We sample each new sparsity level on the line between the current point $(t, s_t)$ and the goal $(t_{\text{final}}, 1)$ of reaching a sparsity level of $1$ at the end of the training as illustrated in Figure~\ref{F:exploration_exploitation_sampling} (middle). This gives the point $(t', s_t + \delta)$, where $\delta = \frac{1 - s_t}{t_{\text{final}} - t} (t' -t)$. To increase exploration, we scale the obtained sparsity level by $U \sim \mathcal{U}(0, U_{\text{max}})$. Additionally, we ensure that the sampled sparsity level does not remove more than $S_{\text{max}} \times 100 \%$ of the remaining parameters such that the jumps in sparsity levels are not too high at the end of the training. This leads to
\begin{equation} \label{E:eaudeqn}
    s_{t'} = s_t + \min \{ \underbrace{\frac{1 - s_t}{t_{\text{final}} - t} (t' - t)}_{\text{linear schedule to sparsity of $1$}} \overbrace{U}^{\mathclap{\substack{\text{stochasticity} \\ \text{injection}}}}, \underbrace{(1 - s_t) S_{\text{max}}}_{\text{geometric speed cap}}\}, \text{where } U \sim \mathcal{U}(0, U_{\text{max}}).
\end{equation}
In practice, setting a sparsity level of $s_t$ is done by updating a binary mask over the weights, where the entries corresponding to the $s_t \times 100 \%$ of the lowest magnitude weights are switched to zero. 

Sampling actions are usually performed using an $\epsilon$-greedy policy computed from the online network~\citep{mnih2015human}. One could consider using the online network with minimal cumulated loss. However, \citet{vincent2024adaqn} argue that it is insufficient because the other networks would learn passively, which is detrimental in the long run~\citep{ostrovski2021difficulty}. Following the recommendations of \citet{vincent2024adaqn}, we sample an online network from a distribution inversely proportional to the cumulated loss as shown in Figure~\ref{F:exploration_exploitation_sampling} (right). Then, an $\epsilon$-greedy policy is built on top of this selected network to foster exploration, as described in Line~\ref{L:action_sampling} in Algorithm~\ref{A:eaudedqn}. 

Overall, this framework is designed to minimize the sum of approximation errors over the training. This motivation is supported by a well-established theoretical result (Theorem $3.4$ from \citet{farahmand2011regularization}) stating that the sum of approximation errors influences a bound on the performance loss, i.e., the distance between the optimal $Q$-function and the $Q$-function related to the greedy policy obtained at the end of the training. As this property is inherited from AdaQN, we refer to \citet{vincent2024adaqn} for further details. In the following, we adapted the presented framework to different algorithms. Each time, we append the name of the algorithm with the prefix "EauDe". As an example, EauDeSAC is an instance of EauDeQN applied to Soft Actor-Critic~(SAC, \citet{pmlr-v80-haarnoja18b}), its pseudo-code is presented in Algorithm~\ref{A:eaudesac}.

\section{Experiments}
We evaluate our approach on $10$ Atari games~\citep{bellemare2013arcade} and $6$ MuJoCo environments~\citep{todorov2012}. We apply EauDeQN on $3$ different algorithms corresponding to $3$ RL settings. We use DQN~\citep{mnih2015human} in an online scenario, Conservative $Q$-Learning~(CQL, \citet{kumar2020conservative}) in an offline scenario, and SAC~\citep{pmlr-v80-haarnoja18b} in an actor-critic setting. In each RL setting, we compare our approach to its dense counterpart and to PolyPruneQN since it is state-of-the-art among pruning methods~\citep{graesser2022state, ceronvalue}. We focus on obtaining returns comparable to those of the dense approach while reaching high final sparsity levels. For that, we report the Inter-Quantile Mean~(IQM, \citet{agarwal2021deep}) of the normalized return and the sparsity levels along with $95\%$ bootstrapped confidence intervals over $5$ seeds for the Atari games and $10$ seeds for the MuJoCo environments. We believe that the number of samples used during training is the main limiting factor for pruning algorithms. This is why we report the number of environment interactions as the $x$-axis, except for the offline experiments where we report the number of batch updates. We use the hyperparameters shared by \citet{ceronvalue} for PolyPruneQN as they demonstrate that their method is also effective on the considered RL setting, i.e., $s_F = 0.95, n = 3, t_{\text{start}} = 0.2 \cdot t_{\text{final}},$ and $t_{\text{end}} = 0.8 \cdot t_{\text{final}},$ where $t_{\text{final}}$ corresponds to the training length. For EauDeQN, we fix $U_{\text{max}} = 3, S_{\text{max}} = 0.01, K = 5$ and $M = 3$ and discuss these values in Section~\ref{S:ablation_study}. The shared hyperparameters are kept fixed across the methods and are reported in Table~\ref{T:atari_parameters} and \ref{T:mujoco_parameters}. We reduced the set of $15$ games selected by \citet{ceronvalue} and \citet{graesser2022state} for their diversity to $10$ games to minimize computational costs. The subset of $10$ games was selected to maintain a wide variety in the magnitude of the normalized return as shown in Figure~\ref{F:game_selection}. Details on experiment settings are shared in Section~\ref{S:appendix}. The individual learning curves for each environment are presented in the supplementary material.

\vspace{-0.3cm}
\subsection{Online Q-Learning} \label{S:online_rl}
\vspace{-0.2cm}
\begin{figure}
    \centering
    \includegraphics[width=\textwidth]{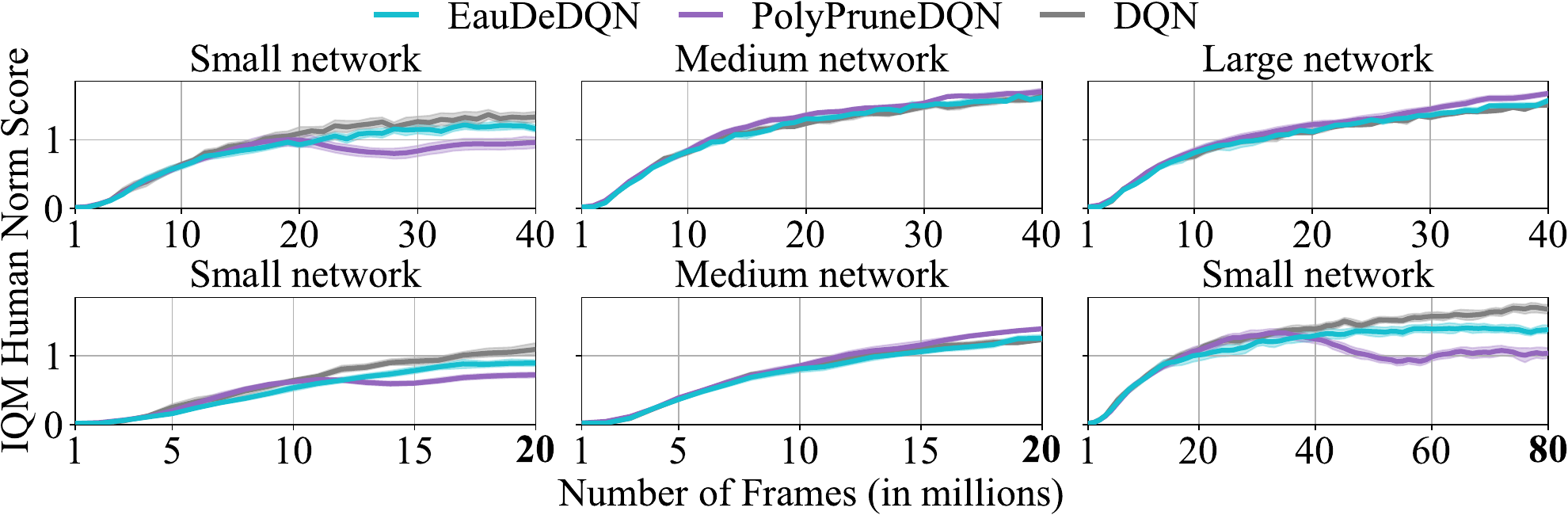}
    \caption{Thanks to its adaptive capability, EauDeDQN performs similarly to its dense counterpart on $10$ \textbf{Atari} games across different network sizes (top row) and training lengths (bottom row). PolyPruneDQN struggles to reach similar returns due to its hard-coded sparsity schedule.}
    \label{F:atari_online_performances}
\end{figure}
\begin{figure}
    \centering
    \includegraphics[width=\textwidth]{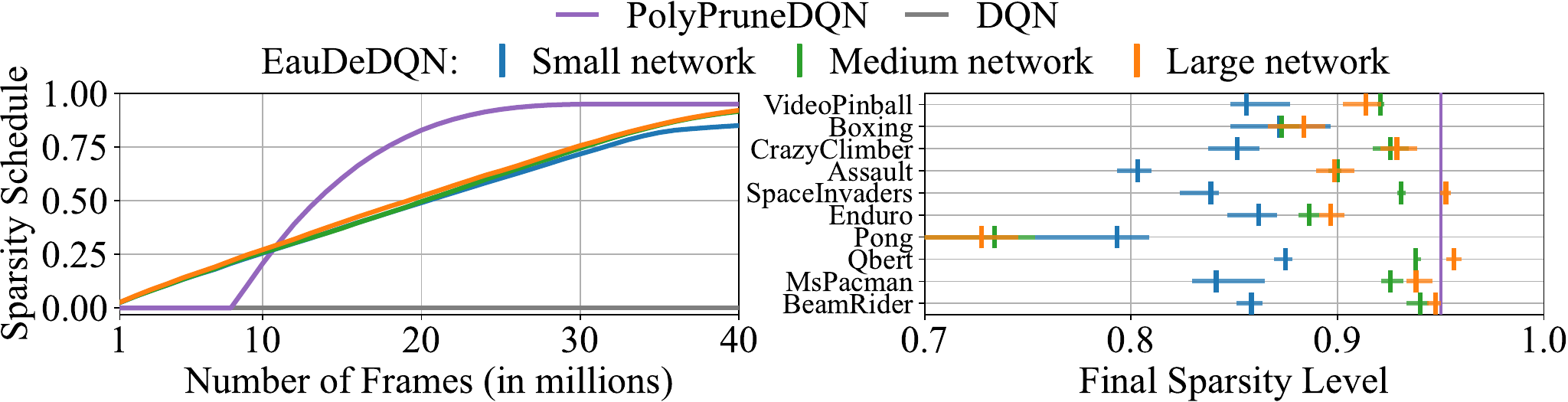}
    \caption{EauDeDQN's sparsity schedule (left) differs across the $3$ tested network sizes. Higher final sparsity levels are reached for larger network sizes at the end of the training (right), showcasing EauDeDQN's adaptivity. The shaded region indicates the variability across seeds.}
    \label{F:atari_online_network_size_sparsity}
\end{figure}
We evaluate EauDeDQN's ability to adapt the sparsity schedule and final sparsity level to different network architectures and training lengths. We make the number of neurons in the first linear layer vary from $32$ (small network) to $512$ (medium network) to $2048$ (large network) while keeping the convolutional layers identical. In Figure~\ref{F:atari_online_performances}, EauDeDQN exhibits a stable behavior across the different network sizes (top row) and training lengths (bottom row). EauDeDQN reaches similar performances compared to its dense counterpart as opposed to PolyPruneDQN, which struggles to obtain high returns with a small network architecture.  

As the representation capacity of the different network architectures is the same (same convolutional layers), one would desire an adaptive pruning algorithm to prune larger networks more, as compared to smaller networks. Figure~\ref{F:atari_online_network_size_sparsity} (left) shows the sparsity schedule obtained by EauDeDQN along with the hard-coded one of PolyPruneDQN. Interestingly, after following a linear curve, the $3$ EauDeDQN's sparsity schedules split into $3$ different curves to end at a final sparsity level that is environment-dependent (Figure~\ref{F:atari_online_network_size_sparsity}, right). We stress that, similarly to PolyPruneDQN, a baseline following a hard-coded linear schedule would also rely on an accurate tuning of its final sparsity level. Notably, except for the game \textit{Pong}, larger final sparsity levels are reached for larger networks, as desired. Figure~\ref{F:training_length_sparsity} (top) exhibits similar behaviors where higher final sparsity levels are discovered when more environment interactions are available.

Could the knowledge about the fact that PolyPruneDQN's medium network performs well with $5\%$ of its weights (Figure~\ref{F:atari_online_performances}, middle), be used to tune PolyPruneDQN's final sparsity level $s_F$ for training the small network using the proportion of the network sizes? As the medium network contains $12.4$ times more weights than the small network (see Table~\ref{T:n_parameters}), the small network should perform well with $62\%$ ($=12.4 \times 5\%$) of its weights. This means that one could set $s_F$ to $0.38$ ($=1 - 0.62$) for training the small network. However, even if PolyPruneDQN would achieve good performances at this final sparsity level, it would be significantly lower than the lowest final sparsity level discovered by EauDeDQN ($0.79$ on \textit{Pong}).

\vspace{-0.1cm}
\subsection{Offline Q-Learning}
\vspace{-0.1cm}
\begin{figure}
    \centering
    \includegraphics[width=\textwidth]{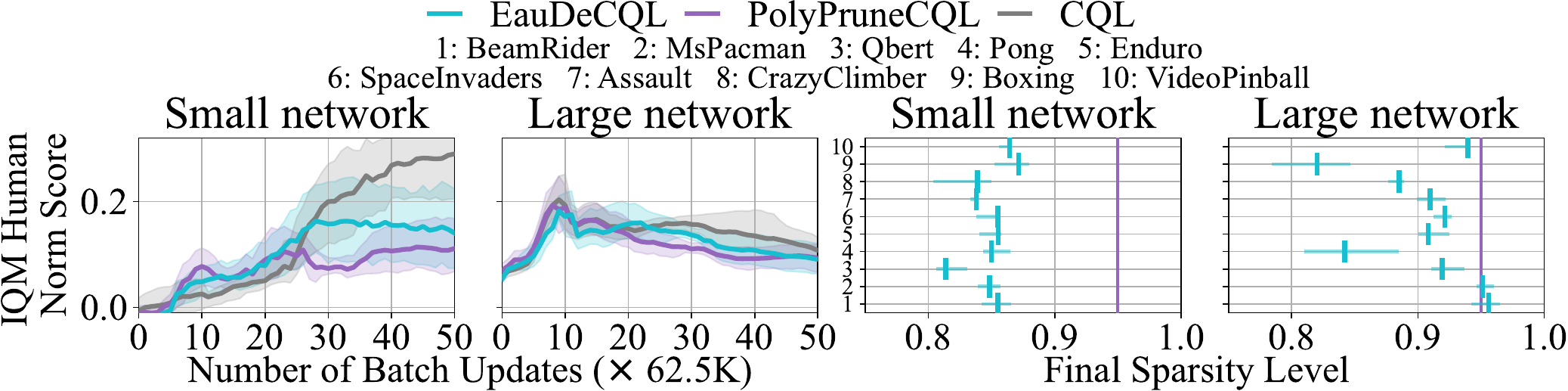}
    \caption{EauDeCQL outperforms PolyPruneCQL when evaluated on $10$ \textbf{Atari} games with a small network and reaches a return similar to the one of CQL with a large network. Importantly, EauDeCQL discovers higher final sparsity levels with a larger network, as desired.}
    \label{F:atari_offline_performances}
    \vspace{-0.7cm}
\end{figure}
EauDeQN is also designed to work offline as it relies on the cumulated loss to select sparsity levels. Therefore, we evaluate the proposed approach on the same set of $10$ Atari games, using an offline dataset that is composed of $5\%$ of the samples collected by a DQN agent during $200$M environment interactions \citep{agarwal2020optimistic}. In Figure~\ref{F:atari_offline_performances} (left), EauDeCQL outperforms PolyPruneCQL for the small network while reaching high sparsity levels, as shown on the right side of the figure. Nonetheless, we note that the confidence intervals overlap and that there is a gap between EauDeCQL and CQL performances. For the larger network, all algorithms reach similar return, with slowly decreasing return over time, as also observed in \citet{ceronvalue}. We attribute this behavior to overfitting as the cumulated losses increase over time (see Figure~\ref{F:atari_offline_cumulated_loss_sparsity}, left). Notably, the sparsity levels reached by EauDeCQL are higher for the larger network, as desired (see Figure~\ref{F:atari_offline_performances}). 

\vspace{-0.1cm}
\subsection{Actor-Critic Method}
\begin{figure}
    \centering
    \includegraphics[width=\textwidth]{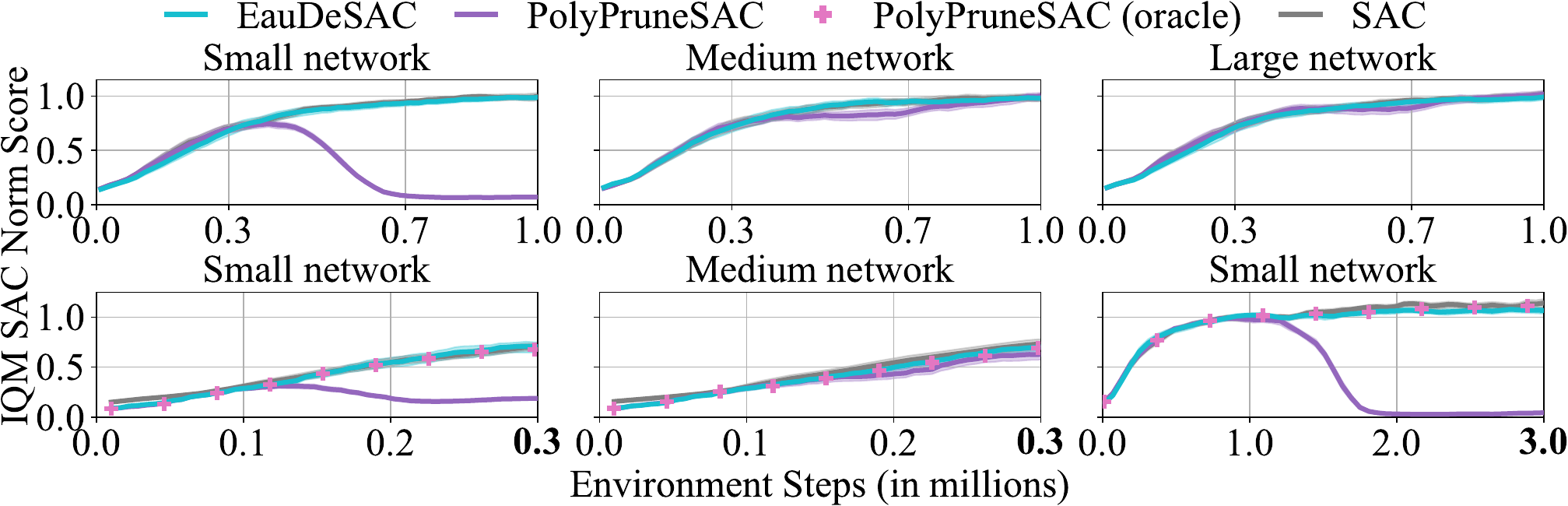}
    \caption{Thanks to its adaptive capability, EauDeSAC performs similarly to its dense counterpart on $6$ \textbf{MuJoCo} games across different network sizes (top row) and training lengths (bottom row). PolyPruneSAC struggles to reach similar returns due to its hard-coded sparsity schedule. PolyPruneSAC (oracle) performs well when the final sparsity is set to the value discovered by EauDeSAC.}
    \label{F:mujoco_performances}
\end{figure}

\vspace{-0.1cm}
We verify that the proposed framework can be used in an actor-critic setting. Similarly to the online Atari experiments in Section~\ref{S:online_rl}, we observe in Figure~\ref{F:mujoco_performances} a stable behavior of EauDeSAC, which yields comparable performances to SAC when the network architecture and the training length vary. On the other hand, PolyPruneSAC suffers when evaluated on small network sizes. The small network corresponds to the commonly used architecture ($256$ neurons for each of the $2$ linear layers \citep{pmlr-v80-haarnoja18b}), the number of neurons per layer is scaled by $5$ for the medium network and by $8$ for the large network. As a sanity check, we verified that the final sparsity levels discovered by EauDeSAC can also be used by PolyPruneSAC to achieve high returns. In Figure~\ref{F:mujoco_performances} (bottom), PolyPruneSAC (oracle) validates this hypothesis by reaching similar performances as SAC and EauDeSAC.

\begin{figure}
    \centering
    \includegraphics[width=\textwidth]{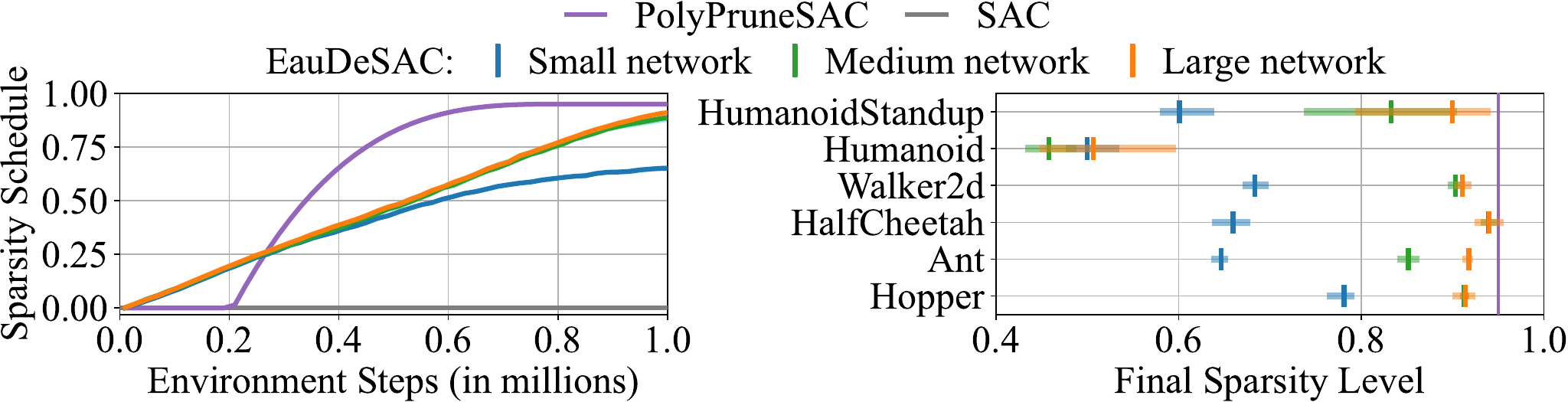}
    \caption{EauDeSAC's sparsity schedule (left) differs across the $3$ tested network sizes. Higher final sparsity levels are reached for larger network sizes at the end of the training (right), showcasing EauDeSAC's adaptivity.}
    \label{F:mujoco_network_size_sparsity}
\end{figure}

Figure~\ref{F:mujoco_network_size_sparsity} shows the sparsity schedules (left) that lead to the final sparsity levels (right). This time, the difference between the $3$ sparsity schedules of EauDeSAC is even more pronounced than for the online Atari experiments. This can be explained by the fact that the differences in scale between the networks are larger than for the Atari experiments (see Table~\ref{T:n_parameters}). Indeed, the small network is $18.8$ times smaller than the medium network and $46.6$ times smaller than the large network. By adaptively selecting the network with the lowest cumulated loss, EauDeSAC filters out the networks with sparsity levels that are too high to fit the regression target. This is why the curve of EauDeSAC's sparsity schedule for the small network is lower than for the larger networks (except for the \textit{Humanoid} environment). Similar conclusions can be drawn for the sparsity schedules obtained with varying training lengths (see Figure~\ref{F:training_length_sparsity}, bottom).

Knowing that PolyPruneSAC's medium network performs well with $5\%$ of its weights can also not be used to tune PolyPruneSAC's final sparsity level for the small network. Indeed, the medium network is $18.8$ times smaller than the small network. This means that the small network could perform well with $94\%$ ($= 18.8 \times 5\%$) of its weights. This leads to a final sparsity level for PolyPruneSAC of $0.06$ ($=1 - 0.94$), which is significantly lower than the lowest sparsity level discovered by EauDeSAC ($0.5$ for Humanoid).

\subsection{Ablation Study} \label{S:ablation_study}
\begin{figure}
    \centering
    \begin{subfigure}{\textwidth}
        \centering
        \includegraphics[width=\textwidth]{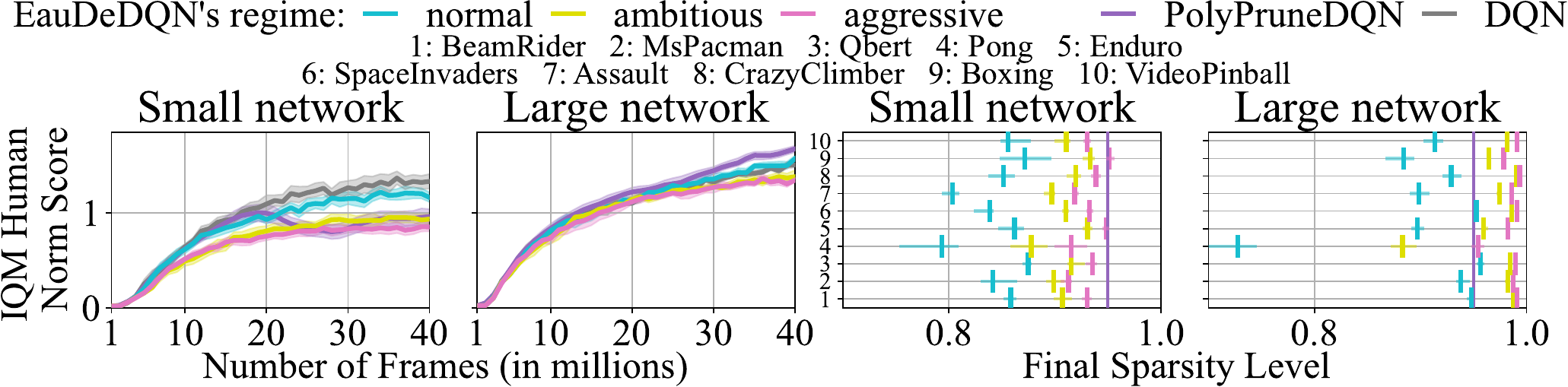}
    \end{subfigure}
    \begin{subfigure}{\textwidth}
        % just to make a space between the two figures
    \end{subfigure}
    \begin{subfigure}{\textwidth}
        \centering
        \includegraphics[width=\textwidth]{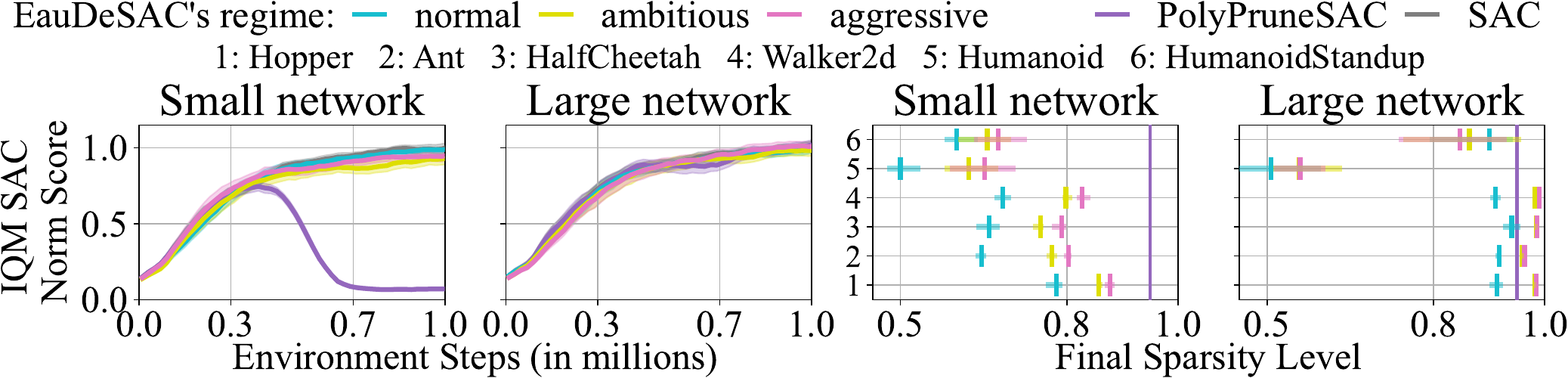}
    \end{subfigure}
    \caption{Evaluation of EauDeDQN on $10$ \textbf{Atari} games (top) and EauDeSAC on $6$ \textbf{MuJoCo} environments (bottom) demonstrating that the tradeoff between high return and high sparsity can be tuned using $U_\text{max}$. For higher values of $U_\text{max}$ (more aggressive regimes), EauDeQN reaches higher final sparsity levels at the cost of a lower return.}
    \label{F:max_noise_performance}
\end{figure}
We now study the sensitivity of EauDeQN to the exploration hyperparameter $U_\text{max}$ introduced in Equation~\ref{E:eaudeqn}. For that, in Figure~\ref{F:max_noise_performance}, we evaluate EauDeDQN (top row) and EauDeSAC (bottom row) with small and large networks, setting $U_\text{max}$ to $3, 10,$ and $30$. This results in a normal, ambitious, and aggressive regime respectively. On both benchmarks, we observe that this hyperparameter offers a tradeoff between high return and high sparsity. As expected, more aggressive regimes constantly yield higher final sparsity levels as higher values of $U_\text{max}$ lead to higher values of sampled sparsity. Across all regimes, we recover the property identified earlier that the final sparsity level for the small networks is lower than for the large network. Figure~\ref{F:max_noise_sparsity} confirms this behavior by showing the sparsity schedules obtained by EauDeDQN (top row) and EauDeSAC (bottom row). Remarkably, with the large network, the aggressive regime reaches final sparsity levels higher than $0.95$ while keeping high performances. We also observe that the aggressive regime is less well suited for the small network on the Atari experiments. Therefore, we recommend increasing the aggressivity of the regime with the network size. In Figure~\ref{F:mujoco_pareto_s_max_k} (left), we compare the Pareto front between sparsity and return of EauDeSAC and PolyPruneSAC. We conclude that EauDeSAC's regime parameter~($U_{\text{max}}$) is easier to tune as setting it too high does not lead to poor performances as opposed to setting PolyPruneSAC's final sparsity level at a high value. Finally, Figure~\ref{F:mujoco_pareto_s_max_k} (right) presents another ablation study on $2$ other hyperparameters~($S_{\text{max}}$ and the population size $K$) showing that EauDeSAC's performance remains stable for a wide range of hyperparameter values.

\vspace{-0.1cm}
\section{Conclusion and Limitations}
We introduced EauDeQN, an algorithm capable of pruning the neural networks' weights at the agent's learning pace. As opposed to current approaches, the final level of sparsity is discovered by the algorithm. These capabilities are achieved by combining DistillQN (also introduced in this work) with AdaQN~\citep{vincent2024adaqn}. We demonstrated that EauDeQN yields high final sparsity levels while keeping performances close to its dense counterpart in a wide variety of problems. 

\textbf{Limitations} EauDeQN requires additional time and memory during training. This is a usual drawback of dense-to-sparse approaches~\citep{graesser2022state}. Importantly, Table~\ref{T:training_time_memory} testifies that training PolyPruneQN with only $2$ different final sparsity levels requires significantly more resources than a single EauDeQN training. Another limitation of our work concerns the actor-critic framework, as it only focuses on pruning the critic. Nonetheless, it is usually the network of the critic that requires a larger amount of parameters~\citep{NEURIPS2020_cceff8fa, kostrikov2021offline, graesser2022state, bhatt2019crossq}. Future work could investigate pruning the actor with a simple hand-designed pruning schedule, as done in \citet{xu2024novel}, while using EauDeQN to prune the critic.

\appendix
\clearpage
\section{Appendix} \label{S:appendix}
Our codebase is written in Jax~\citep{jax2018github} and relies on JaxPruner~\citep{lee2024jaxpruner}. \textbf{The code is available in the supplementary material and will be made open source upon acceptance.} After each exploration step of EauDeQN, we reset the optimizer of the duplicated networks, as advocated by \citet{asadi2023resetting}, while leaving the optimizer of the other networks intact, similarly to PolyPruneQN. 

\begin{wrapfigure}{r}{0.5\textwidth}
    \vspace{-0.5cm}
    \centering
    \includegraphics[width=0.5\textwidth]{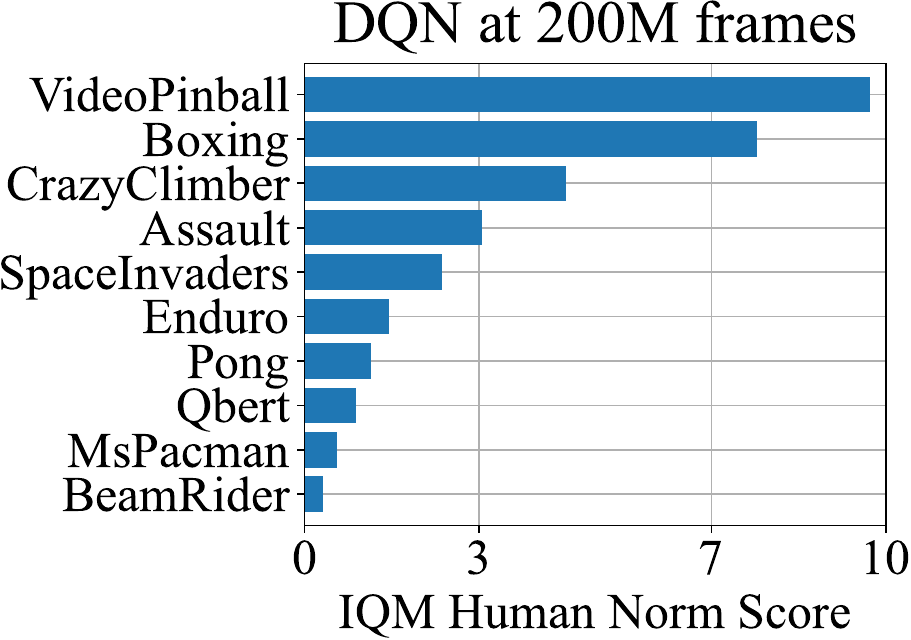}
    \caption{The selected Atari games cover a wide range of normalized returns obtained by DQN after $200$M frames, showcasing their diversity.}
    \label{F:game_selection}
    \vspace{-0.5cm}
\end{wrapfigure}
\textbf{Atari experiment.} We build our codebase on \citet{vincent2025iterated} implementation which follows \citet{castro2018dopamine} standards. Those standards are detailed in \citet{machado2018revisiting}. Namely, we use the \textit{game over} signal to terminate an episode instead of the life signal. The input given to the neural network is a concatenation of $4$ frames in grayscale of dimension $84$ by $84$. To get a new frame, we sample $4$ frames from the Gym environment \citep{brockman2016openai} configured with no frame-skip, and we apply a max pooling operation on the $2$ last grayscale frames. We use sticky actions to make the environment stochastic (with $p = 0.25$). The reported performance is the one obtained during training.

\textbf{MuJoCo experiment.} We build PolyPruneSAC and EauDeSAC on top of SBX~\citep{stable-baselines3}. The agent is evaluated every $10$k environment interaction. While \citet{ceronvalue} apply the hand-designed sparsity schedule on the actor and the critic, in this work, we only prune the critic for PolyPruneSAC and EauDeSAC to remain aligned with the theoretical motivation behind EauDeQN and AdaQN~\citep{vincent2024adaqn}.

\begin{table}[H]
\centering
\caption{Number of parameters of the different network sizes and scaling factor compared to the small network (in parenthesis). Computations were made on the game \textit{SpaceInvaders} and on \textit{Ant}.} \label{T:n_parameters}
\begin{tabular}{ c | c | c | c }
        \toprule
        & Small network & Medium network & Large network \\
        \hline
        Atari & $326\,022$ & $4\,046\,502$ ($\times 12.4$) & $15\,952\,038$ ($\times 48.9$) \\
        \hline
        MuJoCo & $94\,984$ & $1\,785\,608$ ($\times 18.8$) & $4\,429\,832$ ($\times 46.6)$ \\
        \bottomrule
\end{tabular}
\end{table}

\begin{table}[H]
\centering
\caption{While EauDeQN requires additional resources compared to PolyPruneQN, it avoids the need to tune the final sparsity level, which in turn saves resources. Computations are reported for the medium network for every algorithm.}
\label{T:training_time_memory}
\begin{tabular}{ c | c | c | c }
    \toprule
    & EauDeDQN & EauDeCQL & EauDeSAC \\
    & vs PolyPruneDQN & vs PolyPruneCQL & vs PolyPruneSAC \\
    \hline
    Training time & $\times 1,39$ & $\times 1,17$ & $\times 1,08$ \\ 
    \hline
    GPU vRAM usage & $+ 0,73$ Gb & $+ 0,65$ Gb & $+ 0,01$ Gb \\ 
    \hline
    FLOPs for a gradient update & $\times 1.55$ & $\times 1.55$ & $\times 4.03$ \\ 
    \hline
    FLOPs for sampling an action & $\times 1.01$ & (offline) & $\times 1.00$ \\
    \bottomrule
\end{tabular}
\end{table}

\clearpage

\begin{figure}[H]
    \centering
    \begin{subfigure}{\textwidth}
        \centering
        \includegraphics[width=\textwidth]{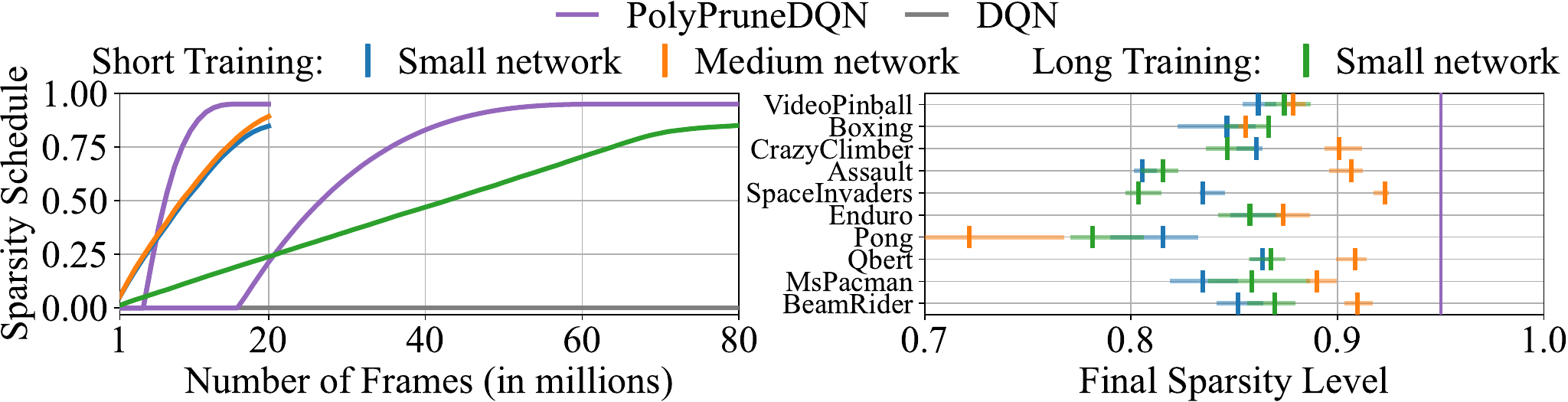}    \end{subfigure}
    \begin{subfigure}{\textwidth}
        \centering
        \vspace{0.1cm}
        \includegraphics[width=\textwidth]{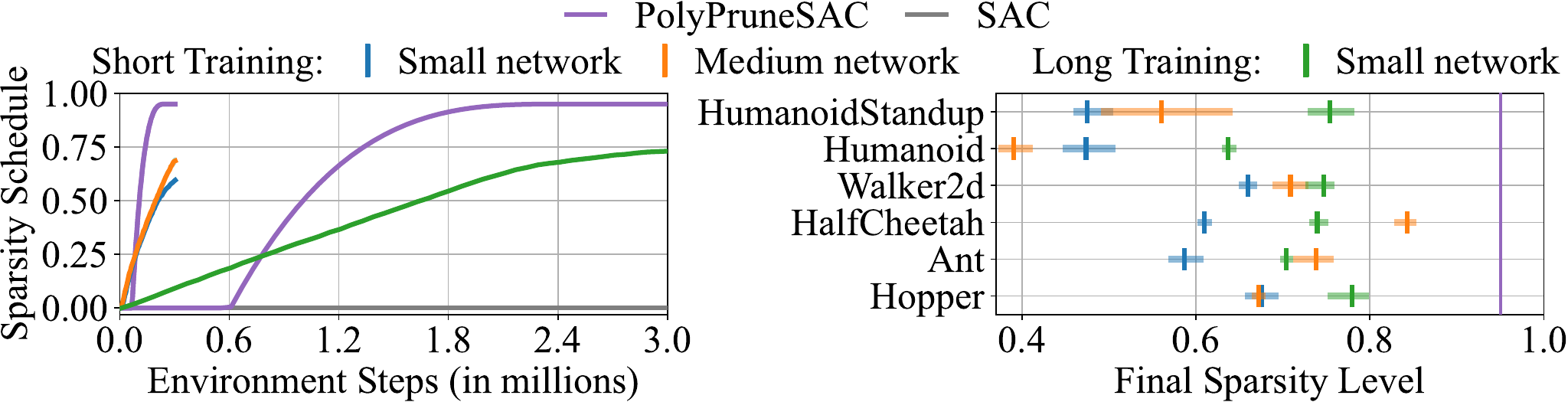}
    \end{subfigure}
    \caption{EauDeDQN (top) and EauDeSAC (bottom) adapt the sparsity schedule to the training length. For small networks, increasing the training length leads to higher final sparsity levels (\textcolor{Blue}{blue} and \textcolor{Green}{green} curves), except for the games \textit{Pong}, \textit{SpaceInvaders}, and \textit{CrazyClimber}. Similarly to Figure~\ref{F:atari_online_network_size_sparsity} and \ref{F:mujoco_network_size_sparsity}, larger networks are pruned at a higher final sparsity level (\textcolor{Blue}{blue} and \textcolor{Orange}{orange} curves), with an exception for \textit{Pong} and \textit{Humanoid}.}
    \label{F:training_length_sparsity}
\end{figure}
\vspace{-0.5cm}
\begin{figure}[H]
    \centering
    \includegraphics[width=\textwidth]{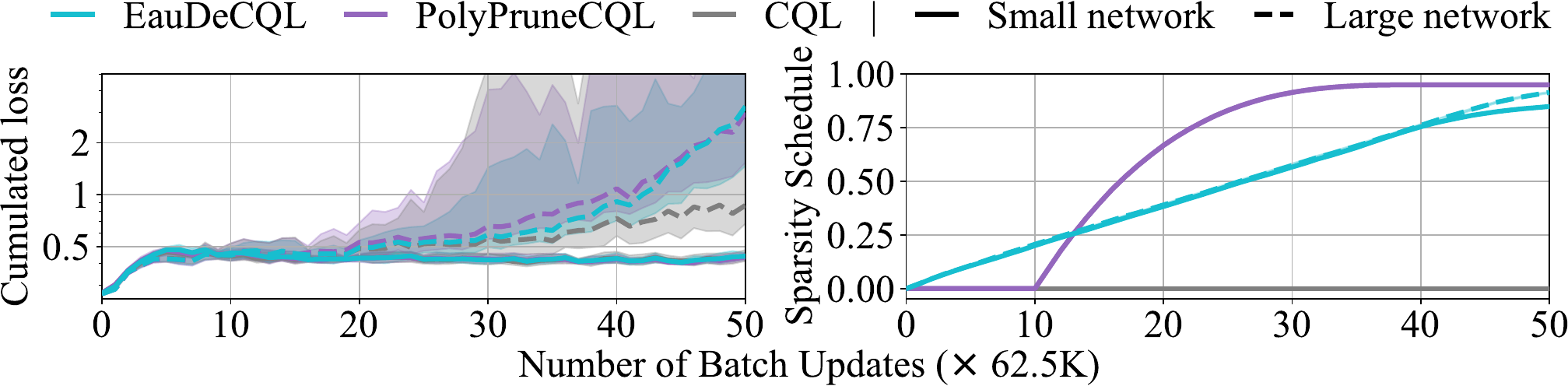}
    \caption{\textbf{Left:} In the offline setting, the larger networks suffer from overfitting as the cumulated losses (reported at every target update and averaged over $T$ updates) increase over time. \textbf{Right:} EauDeCQL adapts the sparsity schedule to the network size. Indeed, sparsity levels are lower for the small network towards the end of the training.}
    \label{F:atari_offline_cumulated_loss_sparsity}
\end{figure}
\vspace{-0.5cm}
\begin{figure}[H]
    \centering
    \includegraphics[width=0.375\textwidth]{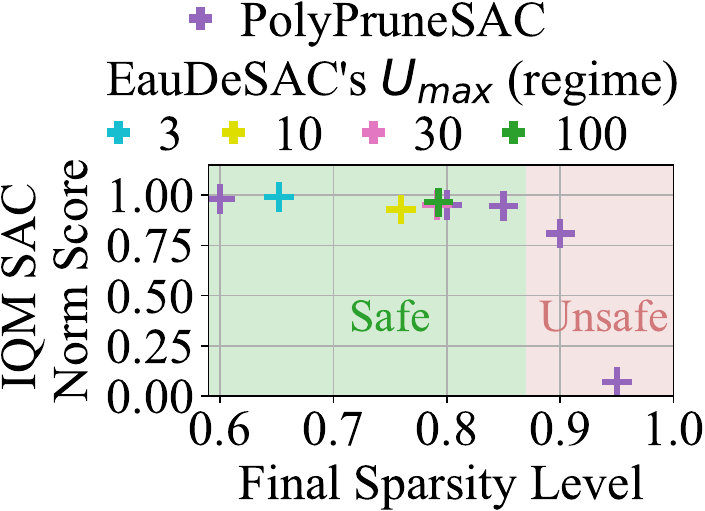}
    \hspace{0.1\textwidth}
    \includegraphics[width=0.5\textwidth]{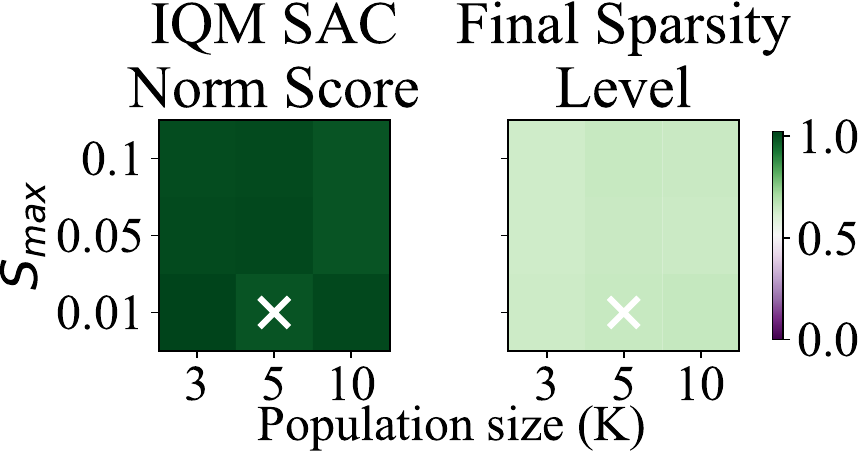}
    \caption{We evaluate EauDeSAC on $6$ MuJoCo environments for $1$M with the small network. \textbf{Left:} PolyPruneSAC requires tuning as its performance depends on its hard-coded final sparsity level. Conversely, EauDeSAC avoids unsafe final sparsity levels by discovering its final sparsity level, therefore requiring only one training to reach a satisfactory outcome. \textbf{Right:} EauDeSAC remains stable across different values of $S_{\text{max}}$ and population size $K$ (see Equation~\ref{E:eaudeqn}), showcasing its robustness w.r.t. hyperparameter changes. The number of subsampled networks $M$ is set to $\left\lceil \frac{K}{2} \right\rceil$. The default hyperparameters of EauDeQN are indicated with a white cross.}
    \label{F:mujoco_pareto_s_max_k}
\end{figure}

\clearpage

\begin{figure}[H]
    \centering
    \begin{subfigure}{\textwidth}
        \centering
        \includegraphics[width=\textwidth]{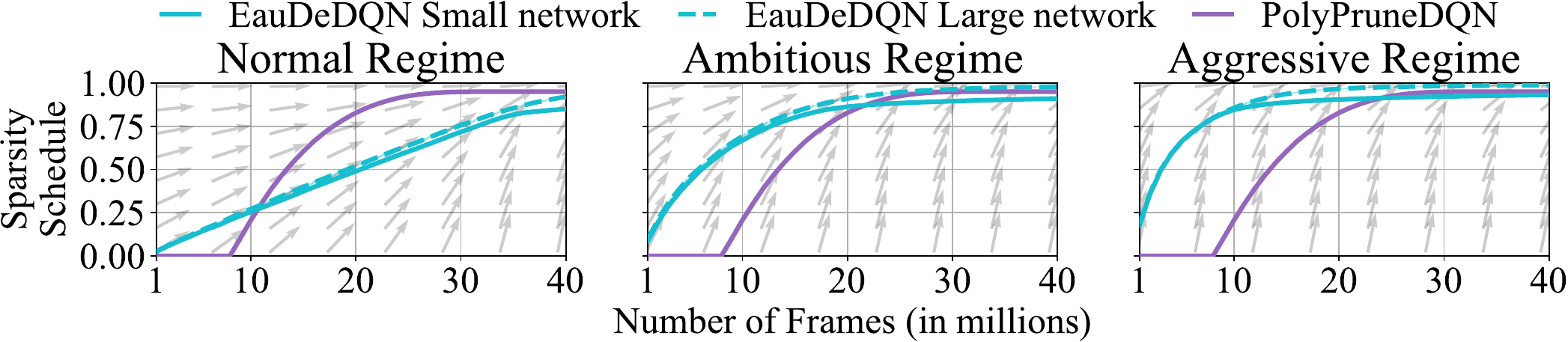}
    \end{subfigure}
    \begin{subfigure}{\textwidth}
        % just to make a space between the two figures
    \end{subfigure}
    \begin{subfigure}{\textwidth}
        \centering
        \includegraphics[width=\textwidth]{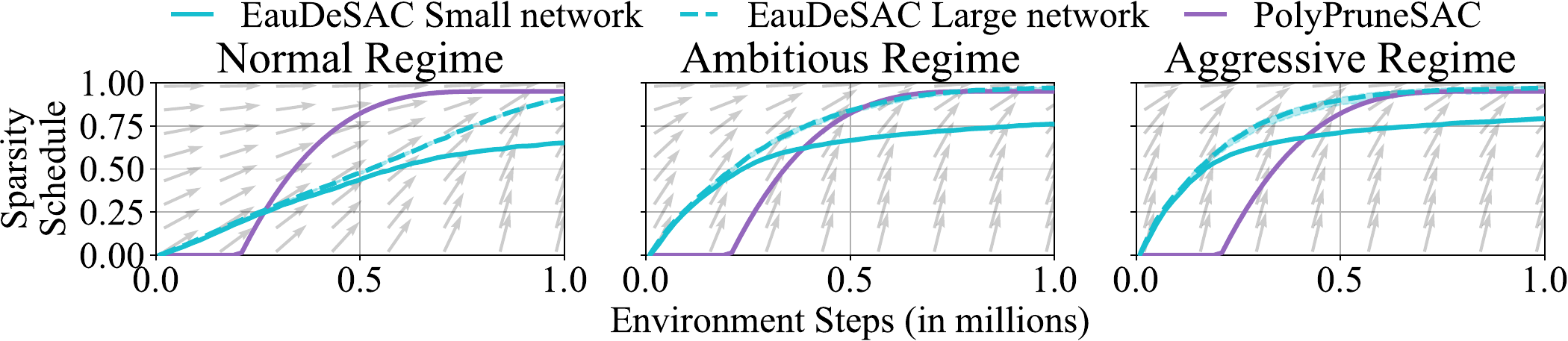}
    \end{subfigure}
    \caption{Sparsity schedules of EauDeDQN on $10$ \textbf{Atari} games (top) and EauDeSAC on $6$ \textbf{MuJoCo} environments (bottom) for different regimes: normal ($U_\text{max} = 3$), ambitious ($U_\text{max} = 10$), and aggressive ($U_\text{max} = 30$). The vector fields in the background show the average direction from which the new sparsities are sampled, similar to Figure~\ref{F:exploration_exploitation_sampling} (middle). As desired, larger networks tend to reach higher sparsity levels. Remarkably, the sparsity levels of the ambitious and aggressive regimes for the large network surpass PolyPruneQN sparsity levels, obtaining higher final sparsity levels while keeping performances high (see Figure~\ref{F:max_noise_performance}).}
    \label{F:max_noise_sparsity}
\end{figure}
\vspace{-0.3cm}
\begin{algorithm}[H]
\caption{Eau De Soft Actor-Critic (EauDeSAC). Modifications to SAC are marked in \textcolor{BlueViolet}{purple}.}
\label{A:eaudesac}
\begin{algorithmic}[1]
\STATE Initialize the policy parameters $\phi$, $2$ \textcolor{BlueViolet}{$\cdot K$} online parameters $(\theta_i^k)_{k = 1}^K$, for $i \in \{1, 2\}$, and an empty replay buffer $\mathcal{D}$. \textcolor{BlueViolet}{For $k=1, .., K$} and $i \in \{1, 2\}$, set the target parameters $\bar{\theta}_i^k \leftarrow \theta_i^k$, and \textcolor{BlueViolet}{the cumulated losses $L_i^k = 0$}. \textcolor{BlueViolet}{Set $\psi_1 = \psi_2 = 0$ the indices to be selected for computing the target.}
\STATE \textbf{repeat}
\begin{ALC@g}
\STATE Take action $a \sim \pi_\phi(\cdot | s)$; Observe reward $r$, next state $s'$; $\mathcal{D} \leftarrow \mathcal{D} \bigcup \{(s, a, r, s')\}$.
\FOR{UTD updates}
\STATE Sample a mini-batch $\mathcal{B} = \{ (s, a, r, s') \}$ from $\mathcal{D}$.
\STATE Compute the \textit{shared} target 
\vspace{-0.3cm}
\begin{equation*}
    y \leftarrow r + \gamma \left( \min_{\bar{\theta} \in \left\{\bar{\theta}_1^{\textcolor{BlueViolet}{\psi_1}}, \bar{\theta}_2^{\textcolor{BlueViolet}{\psi_2}} \right\} } Q_{\bar{\theta}}(s', a') - \alpha \log \pi_\phi(a' | s') \right), \text{where } a' \sim \pi_\phi(\cdot | s' ).
\end{equation*}
\vspace{-0.3cm}
\STATE \textbf{for} \textcolor{BlueViolet}{$k=1, .., K$} and $i=1, 2$ \textbf{do} \textcolor{BlueViolet}{\textit{[in parallel]}}
\begin{ALC@g}
\STATE Compute the loss w.r.t $\theta_i^k$, $\mathcal{L}^{k, i}_{\text{QN}} = \sum_{(s, a, r, s') \in \mathcal{B}} \left( y - Q_{\theta_i^k}(s, a) \right)^2$.
\STATE Update $\theta_i^k$ from $\nabla_{\theta_i^k} \mathcal{L}^{k, i}_{\text{QN}}$, $\bar{\theta}_i^k \leftarrow \tau \theta_i^k + (1 - \tau) \bar{\theta}_i^k$, and \textcolor{BlueViolet}{$L_i^k \leftarrow (1 - \tau) L_i^k + \tau \mathcal{L}^{k, i}_{\text{QN}}$}.
\end{ALC@g}
\STATE \textcolor{BlueViolet}{Set $\psi_i \gets \argmin_{k} L_i^k$, for $i \in \{1, 2\}$.}
\ENDFOR
\STATE \textcolor{BlueViolet}{Set $\psi_i^b \sim \texttt{Choice}(\{1, .., K\}, p=\{\frac{1}{L_i^1}, .., \frac{1}{L_i^K}\})$, for $i \in \{1, 2\}$.}
\STATE Update $\phi$ with gradient ascent using the loss
\vspace{-0.3cm}
\begin{equation*}
    \min_{\theta \in \left\{\theta_1^{\textcolor{BlueViolet}{\psi_1^b}}, \theta_2^{\textcolor{BlueViolet}{\psi_2^b}} \right\} } Q_{\theta}(s, a) - \alpha \log \pi_\phi(a | s), ~~ a \sim \pi_\phi(\cdot | s) \nonumber
\end{equation*}
\vspace{-0.3cm}
\STATE \textbf{every $P$ steps}
\begin{ALC@g}
\STATE \textcolor{BlueViolet}{\texttt{Exploitation:} Select $K$ networks with repetition from the current population using the cumulated losses $L_i^k$. The process is illustrated in Figure~\ref{F:exploration_exploitation_sampling} (left).}
\STATE \textcolor{BlueViolet}{\texttt{Exploration:} Prune the duplicated networks at a sparsity level defined in Equation~\ref{E:eaudeqn}. The process is illustrated in Figure~\ref{F:exploration_exploitation_sampling} (middle).}
\STATE \textcolor{BlueViolet}{Reset $L_i^k \leftarrow 0$, for $k \in \{1, \ldots, K\}$ and $i \in \{1, 2\}$.}
\end{ALC@g}
\end{ALC@g}
\end{algorithmic}
\end{algorithm}

\clearpage
\section*{Acknowledgments}
This work was funded by the German Federal Ministry of Education and Research (BMBF) (Project: 01IS22078). This work was also funded by Hessian.ai through the project ’The Third Wave of Artificial Intelligence – 3AI’ by the Ministry for Science and Arts of the state of Hessen, by the grant “Einrichtung eines Labors des Deutschen Forschungszentrum für Künstliche Intelligenz (DFKI) an der Technischen Universität Darmstadt”, and by the Hessian Ministry of Higher Education, Research, Science and the Arts (HMWK). The authors gratefully acknowledge the scientific support and HPC resources provided by the Erlangen National High Performance Computing Center (NHR@FAU) of the Friedrich-Alexander-Universität Erlangen-Nürnberg (FAU) under the NHR project b187cb. NHR funding is provided by federal and Bavarian state authorities. NHR@FAU hardware is partially funded by the German Research Foundation (DFG) – 440719683.

\subsubsection*{Carbon Impact}
As recommended by \citet{lannelongue2023carbon}, we used GreenAlgorithms \citep{lannelongue2021green} and ML $CO_2$ Impact \citep{lacoste2019quantifying} to compute the carbon emission related to the production of the electricity used for the computations of our experiments. We only consider the energy used to generate the figures presented in this work and ignore the energy used for preliminary studies. The estimations vary between $1.57$ and $1.82$ tonnes of $\text{CO}_2$ equivalent. As a reminder, the Intergovernmental Panel on Climate Change advocates a carbon budget of $2$ tonnes of $\text{CO}_2$ equivalent per year per person.

\bibliography{main}

\begin{thebibliography}{53}
\providecommand{\natexlab}[1]{#1}
\providecommand{\url}[1]{\texttt{#1}}
\expandafter\ifx\csname urlstyle\endcsname\relax
  \providecommand{\doi}[1]{DOI: #1}\else
  \providecommand{\doi}{DOI: \begingroup \urlstyle{rm}\Url}\fi

\bibitem[Agarwal et~al.(2020)Agarwal, Schuurmans, and Norouzi]{agarwal2020optimistic}
Rishabh Agarwal, Dale Schuurmans, and Mohammad Norouzi.
\newblock An optimistic perspective on offline reinforcement learning.
\newblock In \emph{International Conference on Machine Learning}, 2020.

\bibitem[Agarwal et~al.(2021)Agarwal, Schwarzer, Castro, Courville, and Bellemare]{agarwal2021deep}
Rishabh Agarwal, Max Schwarzer, Pablo~Samuel Castro, Aaron Courville, and Marc Bellemare.
\newblock Deep reinforcement learning at the edge of the statistical precipice.
\newblock In \emph{Advances in Neural Information Processing Systems}, 2021.

\bibitem[Arnob et~al.(2021)Arnob, Ohib, Plis, and Precup]{arnob2021single}
Samin~Yeasar Arnob, Riyasat Ohib, Sergey Plis, and Doina Precup.
\newblock Single-shot pruning for offline reinforcement learning.
\newblock \emph{Neurips Workshop on Offline Reinforcement Learning}, 2021.

\bibitem[Asadi et~al.(2023)Asadi, Fakoor, and Sabach]{asadi2023resetting}
Kavosh Asadi, Rasool Fakoor, and Shoham Sabach.
\newblock Resetting the optimizer in deep {RL}: An empirical study.
\newblock In \emph{Advances in Neural Information Processing Systems}, 2023.

\bibitem[Bellemare et~al.(2013)Bellemare, Naddaf, Veness, and Bowling]{bellemare2013arcade}
Marc~G Bellemare, Yavar Naddaf, Joel Veness, and Michael Bowling.
\newblock The arcade learning environment: An evaluation platform for general agents.
\newblock \emph{Journal of Artificial Intelligence Research}, 2013.

\bibitem[Bhatt et~al.(2024)Bhatt, Palenicek, Belousov, Argus, Amiranashvili, Brox, and Peters]{bhatt2019crossq}
Aditya Bhatt, Daniel Palenicek, Boris Belousov, Max Argus, Artemij Amiranashvili, Thomas Brox, and Jan Peters.
\newblock Crossq: Batch normalization in deep reinforcement learning for greater sample efficiency and simplicity.
\newblock \emph{International Conference on Learning Representations}, 2024.

\bibitem[Bradbury et~al.(2018)Bradbury, Frostig, Hawkins, Johnson, Leary, Maclaurin, Necula, Paszke, Vander{P}las, Wanderman-{M}ilne, and Zhang]{jax2018github}
James Bradbury, Roy Frostig, Peter Hawkins, Matthew~James Johnson, Chris Leary, Dougal Maclaurin, George Necula, Adam Paszke, Jake Vander{P}las, Skye Wanderman-{M}ilne, and Qiao Zhang.
\newblock \emph{{JAX}: composable transformations of {P}ython+{N}um{P}y programs}, 2018.

\bibitem[Brockman et~al.(2016)Brockman, Cheung, Pettersson, Schneider, Schulman, Tang, and Zaremba]{brockman2016openai}
Greg Brockman, Vicki Cheung, Ludwig Pettersson, Jonas Schneider, John Schulman, Jie Tang, and Wojciech Zaremba.
\newblock Openai gym.
\newblock \emph{arXiv preprint arXiv:1606.01540}, 2016.

\bibitem[Castro et~al.(2018)Castro, Moitra, Gelada, Kumar, and Bellemare]{castro2018dopamine}
Pablo~Samuel Castro, Subhodeep Moitra, Carles Gelada, Saurabh Kumar, and Marc~G Bellemare.
\newblock Dopamine: A research framework for deep reinforcement learning.
\newblock \emph{arXiv preprint arXiv:1812.06110}, 2018.

\bibitem[Ceron et~al.(2024)Ceron, Courville, and Castro]{ceronvalue}
Johan Samir~Obando Ceron, Aaron Courville, and Pablo~Samuel Castro.
\newblock In value-based deep reinforcement learning, a pruned network is a good network.
\newblock In \emph{International Conference on Machine Learning}, 2024.

\bibitem[Ernst et~al.(2005)Ernst, Geurts, and Wehenkel]{ernst05a}
Damien Ernst, Pierre Geurts, and Louis Wehenkel.
\newblock Tree-based batch mode reinforcement learning.
\newblock \emph{Journal of Machine Learning Research}, 2005.

\bibitem[Espeholt et~al.(2018)Espeholt, Soyer, Munos, Simonyan, Mnih, Ward, Doron, Firoiu, Harley, Dunning, et~al.]{espeholt2018impala}
Lasse Espeholt, Hubert Soyer, Remi Munos, Karen Simonyan, Vlad Mnih, Tom Ward, Yotam Doron, Vlad Firoiu, Tim Harley, Iain Dunning, et~al.
\newblock Impala: Scalable distributed deep-rl with importance weighted actor-learner architectures.
\newblock In \emph{International Conference on Machine Learning}, 2018.

\bibitem[Evci et~al.(2019)Evci, Pedregosa, Gomez, and Elsen]{evci2019difficulty}
Utku Evci, Fabian Pedregosa, Aidan Gomez, and Erich Elsen.
\newblock The difficulty of training sparse neural networks.
\newblock In \emph{ICML Workshop on Identifying and Understanding Deep Learning Phenomena}, 2019.

\bibitem[Evci et~al.(2020)Evci, Gale, Menick, Castro, and Elsen]{evci2020rigging}
Utku Evci, Trevor Gale, Jacob Menick, Pablo~Samuel Castro, and Erich Elsen.
\newblock Rigging the lottery: Making all tickets winners.
\newblock In \emph{International Conference on Machine Learning}, 2020.

\bibitem[Farahmand(2011)]{farahmand2011regularization}
Amir-massoud Farahmand.
\newblock \emph{Regularization in reinforcement learning}.
\newblock PhD thesis, University of Alberta, 2011.

\bibitem[Franke et~al.(2021)Franke, Koehler, Biedenkapp, and Hutter]{frankesample}
J{\"o}rg~KH Franke, Gregor Koehler, Andr{\'e} Biedenkapp, and Frank Hutter.
\newblock Sample-efficient automated deep reinforcement learning.
\newblock In \emph{International Conference on Learning Representations}, 2021.

\bibitem[Frankle \& Carbin(2018)Frankle and Carbin]{frankle2018lottery}
Jonathan Frankle and Michael Carbin.
\newblock The lottery ticket hypothesis: Finding sparse, trainable neural networks.
\newblock In \emph{International Conference on Learning Representations}, 2018.

\bibitem[Graesser et~al.(2022)Graesser, Evci, Elsen, and Castro]{graesser2022state}
Laura Graesser, Utku Evci, Erich Elsen, and Pablo~Samuel Castro.
\newblock The state of sparse training in deep reinforcement learning.
\newblock In \emph{International Conference on Machine Learning}, 2022.

\bibitem[Grooten et~al.(2023)Grooten, Sokar, Dohare, Mocanu, Taylor, Pechenizkiy, and Mocanu]{grooten2023automatic}
Bram Grooten, Ghada Sokar, Shibhansh Dohare, Elena Mocanu, Matthew~E Taylor, Mykola Pechenizkiy, and Decebal~Constantin Mocanu.
\newblock Automatic noise filtering with dynamic sparse training in deep reinforcement learning.
\newblock In \emph{International Conference on Autonomous Agents and Multiagent System}, 2023.

\bibitem[Haarnoja et~al.(2018)Haarnoja, Zhou, Abbeel, and Levine]{pmlr-v80-haarnoja18b}
Tuomas Haarnoja, Aurick Zhou, Pieter Abbeel, and Sergey Levine.
\newblock Soft actor-critic: Off-policy maximum entropy deep reinforcement learning with a stochastic actor.
\newblock In \emph{International Conference on Machine Learning}, 2018.

\bibitem[Han et~al.(2015)Han, Mao, and Dally]{han2015deep}
Song Han, Huizi Mao, and William~J Dally.
\newblock Deep compression: Compressing deep neural networks with pruning, trained quantization and huffman coding.
\newblock \emph{International Conference on Learning Representations}, 2015.

\bibitem[Henderson et~al.(2018)Henderson, Islam, Bachman, Pineau, Precup, and Meger]{henderson2018deep}
Peter Henderson, Riashat Islam, Philip Bachman, Joelle Pineau, Doina Precup, and David Meger.
\newblock Deep reinforcement learning that matters.
\newblock In \emph{Association for the Advancement of Artificial Intelligence}, 2018.

\bibitem[Kostrikov et~al.(2021)Kostrikov, Fergus, Tompson, and Nachum]{kostrikov2021offline}
Ilya Kostrikov, Rob Fergus, Jonathan Tompson, and Ofir Nachum.
\newblock Offline reinforcement learning with fisher divergence critic regularization.
\newblock In \emph{International Conference on Machine Learning}, 2021.

\bibitem[Kumar et~al.(2020)Kumar, Zhou, Tucker, and Levine]{kumar2020conservative}
Aviral Kumar, Aurick Zhou, George Tucker, and Sergey Levine.
\newblock Conservative q-learning for offline reinforcement learning.
\newblock \emph{Advances in Neural Information Processing Systems}, 2020.

\bibitem[Lacoste et~al.(2019)Lacoste, Luccioni, Schmidt, and Dandres]{lacoste2019quantifying}
Alexandre Lacoste, Alexandra Luccioni, Victor Schmidt, and Thomas Dandres.
\newblock Quantifying the carbon emissions of machine learning.
\newblock \emph{arXiv preprint arXiv:1910.09700}, 2019.

\bibitem[Lannelongue \& Inouye(2023)Lannelongue and Inouye]{lannelongue2023carbon}
Lo{\"\i}c Lannelongue and Michael Inouye.
\newblock Carbon footprint estimation for computational research.
\newblock \emph{Nature Reviews Methods Primers}, 2023.

\bibitem[Lannelongue et~al.(2021)Lannelongue, Grealey, and Inouye]{lannelongue2021green}
Lo{\"\i}c Lannelongue, Jason Grealey, and Michael Inouye.
\newblock Green algorithms: quantifying the carbon footprint of computation.
\newblock \emph{Advanced Science}, 2021.

\bibitem[Lee et~al.(2024)Lee, Park, Mitchell, Pilault, Ceron, Kim, Lee, Frantar, Long, Yazdanbakhsh, et~al.]{lee2024jaxpruner}
Joo~Hyung Lee, Wonpyo Park, Nicole~Elyse Mitchell, Jonathan Pilault, Johan Samir~Obando Ceron, Han-Byul Kim, Namhoon Lee, Elias Frantar, Yun Long, Amir Yazdanbakhsh, et~al.
\newblock Jaxpruner: A concise library for sparsity research.
\newblock In \emph{Conference on Parsimony and Learning}, 2024.

\bibitem[Liu et~al.(2020)Liu, Xu, Shi, Cheung, and So]{liu2020dynamic}
Junjie Liu, Zhe Xu, Runbin Shi, Ray~CC Cheung, and Hayden~KH So.
\newblock Dynamic sparse training: Find efficient sparse network from scratch with trainable masked layers.
\newblock \emph{International Conference on Learning Representations}, 2020.

\bibitem[Liu et~al.(2019)Liu, Sun, Zhou, Huang, and Darrell]{liurethinking}
Zhuang Liu, Mingjie Sun, Tinghui Zhou, Gao Huang, and Trevor Darrell.
\newblock Rethinking the value of network pruning.
\newblock In \emph{International Conference on Learning Representations}, 2019.

\bibitem[Livne \& Cohen(2020)Livne and Cohen]{livne2020pops}
Dor Livne and Kobi Cohen.
\newblock Pops: Policy pruning and shrinking for deep reinforcement learning.
\newblock \emph{IEEE Journal of Selected Topics in Signal Processing}, 2020.

\bibitem[Machado et~al.(2018)Machado, Bellemare, Talvitie, Veness, Hausknecht, and Bowling]{machado2018revisiting}
Marlos~C Machado, Marc~G Bellemare, Erik Talvitie, Joel Veness, Matthew Hausknecht, and Michael Bowling.
\newblock Revisiting the arcade learning environment: Evaluation protocols and open problems for general agents.
\newblock \emph{Journal of Artificial Intelligence Research}, 2018.

\bibitem[Miller et~al.(1995)Miller, Goldberg, et~al.]{miller1995genetic}
Brad~L Miller, David~E Goldberg, et~al.
\newblock Genetic algorithms, tournament selection, and the effects of noise.
\newblock \emph{Complex systems}, 1995.

\bibitem[Mnih et~al.(2015)Mnih, Kavukcuoglu, Silver, Rusu, Veness, Bellemare, Graves, Riedmiller, Fidjeland, Ostrovski, et~al.]{mnih2015human}
Volodymyr Mnih, Koray Kavukcuoglu, David Silver, Andrei~A Rusu, Joel Veness, Marc~G Bellemare, Alex Graves, Martin Riedmiller, Andreas~K Fidjeland, Georg Ostrovski, et~al.
\newblock Human-level control through deep reinforcement learning.
\newblock \emph{Nature}, 2015.

\bibitem[Mocanu et~al.(2018)Mocanu, Mocanu, Stone, Nguyen, Gibescu, and Liotta]{mocanu2018scalable}
Decebal~Constantin Mocanu, Elena Mocanu, Peter Stone, Phuong~H Nguyen, Madeleine Gibescu, and Antonio Liotta.
\newblock Scalable training of artificial neural networks with adaptive sparse connectivity inspired by network science.
\newblock \emph{Nature Communications}, 2018.

\bibitem[Molchanov et~al.(2017)Molchanov, Ashukha, and Vetrov]{molchanov2017variational}
Dmitry Molchanov, Arsenii Ashukha, and Dmitry Vetrov.
\newblock Variational dropout sparsifies deep neural networks.
\newblock In \emph{International Conference on Machine Learning}, 2017.

\bibitem[Nauman et~al.(2024)Nauman, Ostaszewski, Jankowski, Mi{\l}o{\'s}, and Cygan]{nauman2024bigger}
Michal Nauman, Mateusz Ostaszewski, Krzysztof Jankowski, Piotr Mi{\l}o{\'s}, and Marek Cygan.
\newblock Bigger, regularized, optimistic: scaling for compute and sample-efficient continuous control.
\newblock \emph{Advances in Neural Information Processing Systems}, 2024.

\bibitem[Ostrovski et~al.(2021)Ostrovski, Castro, and Dabney]{ostrovski2021difficulty}
Georg Ostrovski, Pablo~Samuel Castro, and Will Dabney.
\newblock The difficulty of passive learning in deep reinforcement learning.
\newblock \emph{Advances in Neural Information Processing Systems}, 34:\penalty0 23283--23295, 2021.

\bibitem[Ota et~al.(2024)Ota, Jha, and Kanezaki]{ota2024training}
Kei Ota, Devesh~K Jha, and Asako Kanezaki.
\newblock Training larger networks for deep reinforcement learning.
\newblock \emph{Machine Learning}, 2024.

\bibitem[Raffin et~al.(2021)Raffin, Hill, Gleave, Kanervisto, Ernestus, and Dormann]{stable-baselines3}
Antonin Raffin, Ashley Hill, Adam Gleave, Anssi Kanervisto, Maximilian Ernestus, and Noah Dormann.
\newblock Stable-baselines3: Reliable reinforcement learning implementations.
\newblock \emph{Journal of Machine Learning Research}, 2021.

\bibitem[Schmitt et~al.(2018)Schmitt, Hudson, Zidek, Osindero, Doersch, Czarnecki, Leibo, Kuttler, Zisserman, Simonyan, et~al.]{schmitt2018kickstarting}
Simon Schmitt, Jonathan~J Hudson, Augustin Zidek, Simon Osindero, Carl Doersch, Wojciech~M Czarnecki, Joel~Z Leibo, Heinrich Kuttler, Andrew Zisserman, Karen Simonyan, et~al.
\newblock Kickstarting deep reinforcement learning.
\newblock \emph{NeurIPS Workshop on Deep Reinforcement Learning}, 2018.

\bibitem[Schwarzer et~al.(2023)Schwarzer, Ceron, Courville, Bellemare, Agarwal, and Castro]{schwarzer2023bigger}
Max Schwarzer, Johan Samir~Obando Ceron, Aaron Courville, Marc~G Bellemare, Rishabh Agarwal, and Pablo~Samuel Castro.
\newblock Bigger, better, faster: Human-level atari with human-level efficiency.
\newblock In \emph{International Conference on Machine Learning}, 2023.

\bibitem[Sokar et~al.(2021)Sokar, Mocanu, Mocanu, Pechenizkiy, and Stone]{sokar2021dynamic}
Ghada Sokar, Elena Mocanu, Decebal~Constantin Mocanu, Mykola Pechenizkiy, and Peter Stone.
\newblock Dynamic sparse training for deep reinforcement learning.
\newblock \emph{International Joint Conference on Artificial Intelligence}, 2021.

\bibitem[Sutton \& Barto(1998)Sutton and Barto]{sutton1998rli}
Richard Sutton and Andrew Barto.
\newblock \emph{Reinforcement learning: An introduction}.
\newblock MIT Press, 1998.

\bibitem[Tan et~al.(2023)Tan, Hu, Pan, Huang, and Huang]{tanrlx2}
Yiqin Tan, Pihe Hu, Ling Pan, Jiatai Huang, and Longbo Huang.
\newblock Rlx2: Training a sparse deep reinforcement learning model from scratch.
\newblock In \emph{International Conference on Learning Representations}, 2023.

\bibitem[Todorov et~al.(2012)Todorov, Erez, and Tassa]{todorov2012}
Emanuel Todorov, Tom Erez, and Yuval Tassa.
\newblock Mujoco: A physics engine for model-based control.
\newblock In \emph{International Conference on Intelligent Robots and Systems}, 2012.

\bibitem[Vincent et~al.(2025{\natexlab{a}})Vincent, Palenicek, Belousov, Peters, and D'Eramo]{vincent2025iterated}
Th{\'e}o Vincent, Daniel Palenicek, Boris Belousov, Jan Peters, and Carlo D'Eramo.
\newblock Iterated $ q $-network: Beyond one-step bellman updates in deep reinforcement learning.
\newblock \emph{Transactions on Machine Learning Research}, 2025{\natexlab{a}}.

\bibitem[Vincent et~al.(2025{\natexlab{b}})Vincent, Wahren, Peters, Belousov, and D'Eramo]{vincent2024adaqn}
Th{\'e}o Vincent, Fabian Wahren, Jan Peters, Boris Belousov, and Carlo D'Eramo.
\newblock Adaptive $ q $-network: On-the-fly target selection for deep reinforcement learning.
\newblock In \emph{International Conference on Learning Representations}, 2025{\natexlab{b}}.

\bibitem[Xu et~al.(2024)Xu, Chen, and Wang]{xu2024novel}
Meng Xu, Xinhong Chen, and Jianping Wang.
\newblock A novel topology adaptation strategy for dynamic sparse training in deep reinforcement learning.
\newblock \emph{IEEE Transactions on Neural Networks and Learning Systems}, 2024.

\bibitem[Yu et~al.(2019)Yu, Edunov, Tian, and Morcos]{yu2019playing}
Haonan Yu, Sergey Edunov, Yuandong Tian, and Ari~S Morcos.
\newblock Playing the lottery with rewards and multiple languages: lottery tickets in rl and nlp.
\newblock \emph{International Conference on Learning Representations}, 2019.

\bibitem[Zhang et~al.(2019)Zhang, He, and Li]{zhang2019accelerating}
Hongjie Zhang, Zhuocheng He, and Jing Li.
\newblock Accelerating the deep reinforcement learning with neural network compression.
\newblock In \emph{International Joint Conference on Neural Networks}, 2019.

\bibitem[Zhou et~al.(2020)Zhou, Li, Yang, Wang, and Hospedales]{NEURIPS2020_cceff8fa}
Wei Zhou, Yiying Li, Yongxin Yang, Huaimin Wang, and Timothy Hospedales.
\newblock Online meta-critic learning for off-policy actor-critic methods.
\newblock In \emph{Advances in Neural Information Processing Systems}, 2020.

\bibitem[Zhu \& Gupta(2018)Zhu and Gupta]{zhu2018prune}
Michael~H Zhu and Suyog Gupta.
\newblock To prune, or not to prune: Exploring the efficacy of pruning for model compression.
\newblock In \emph{ICLR Workshop}, 2018.

\end{thebibliography}
\bibliographystyle{rlj}

\beginSupplementaryMaterials
\begin{figure}[H]
    \centering
    \includegraphics[width=\textwidth]{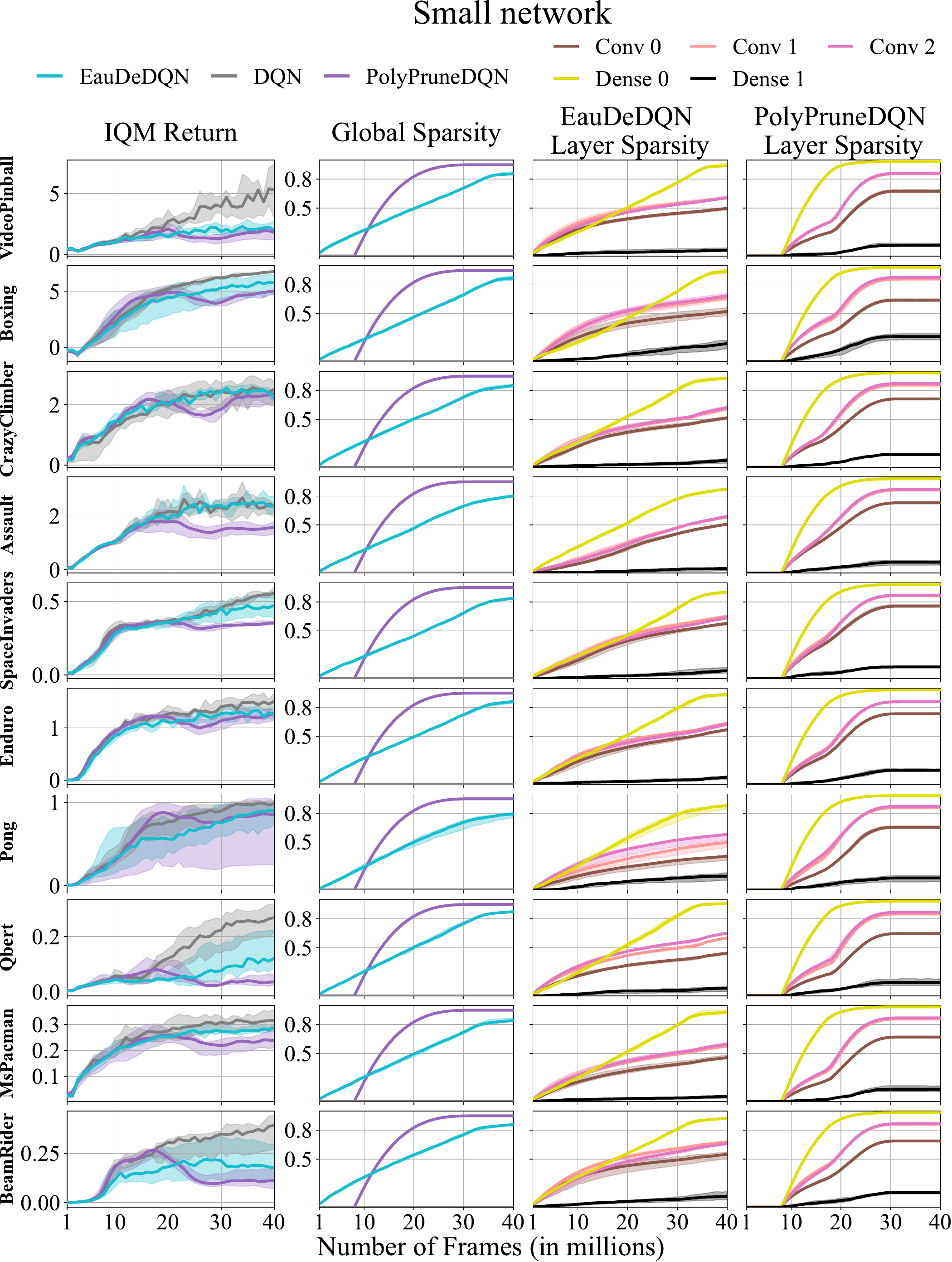}
    \caption{\textbf{Online Atari:} Per game metrics for the experiment on the small network. The aggregated performances are available in Figure~\ref{F:atari_online_performances} (top, left).}
\end{figure}
\begin{figure}[H]
    \centering
    \includegraphics[width=\textwidth]{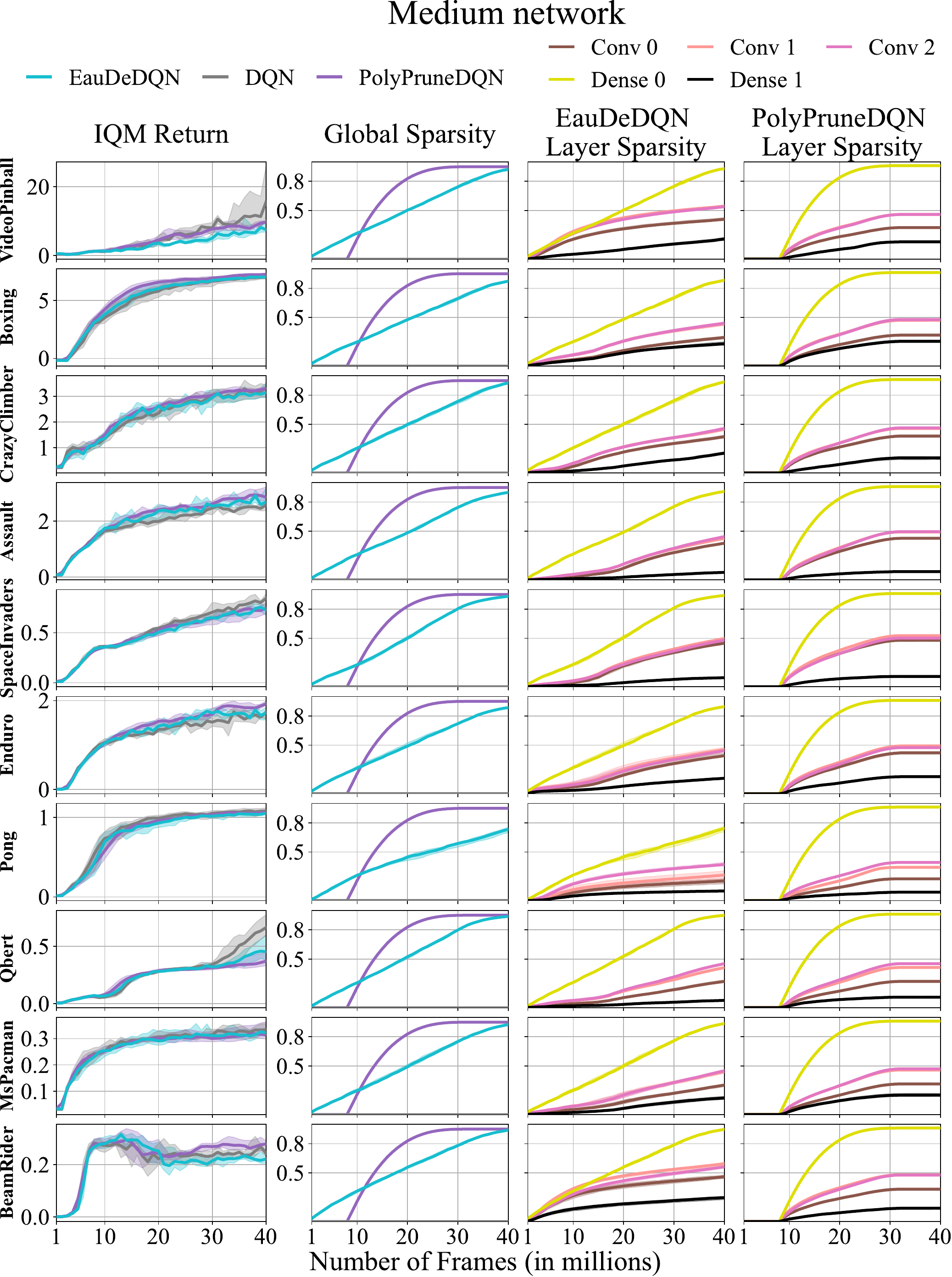}
    \caption{\textbf{Online Atari:} Per game metrics for the experiment on the medium network. The aggregated performances are available in Figure~\ref{F:atari_online_performances} (top, middle).}
\end{figure}
\begin{figure}[H]
    \centering
    \includegraphics[width=\textwidth]{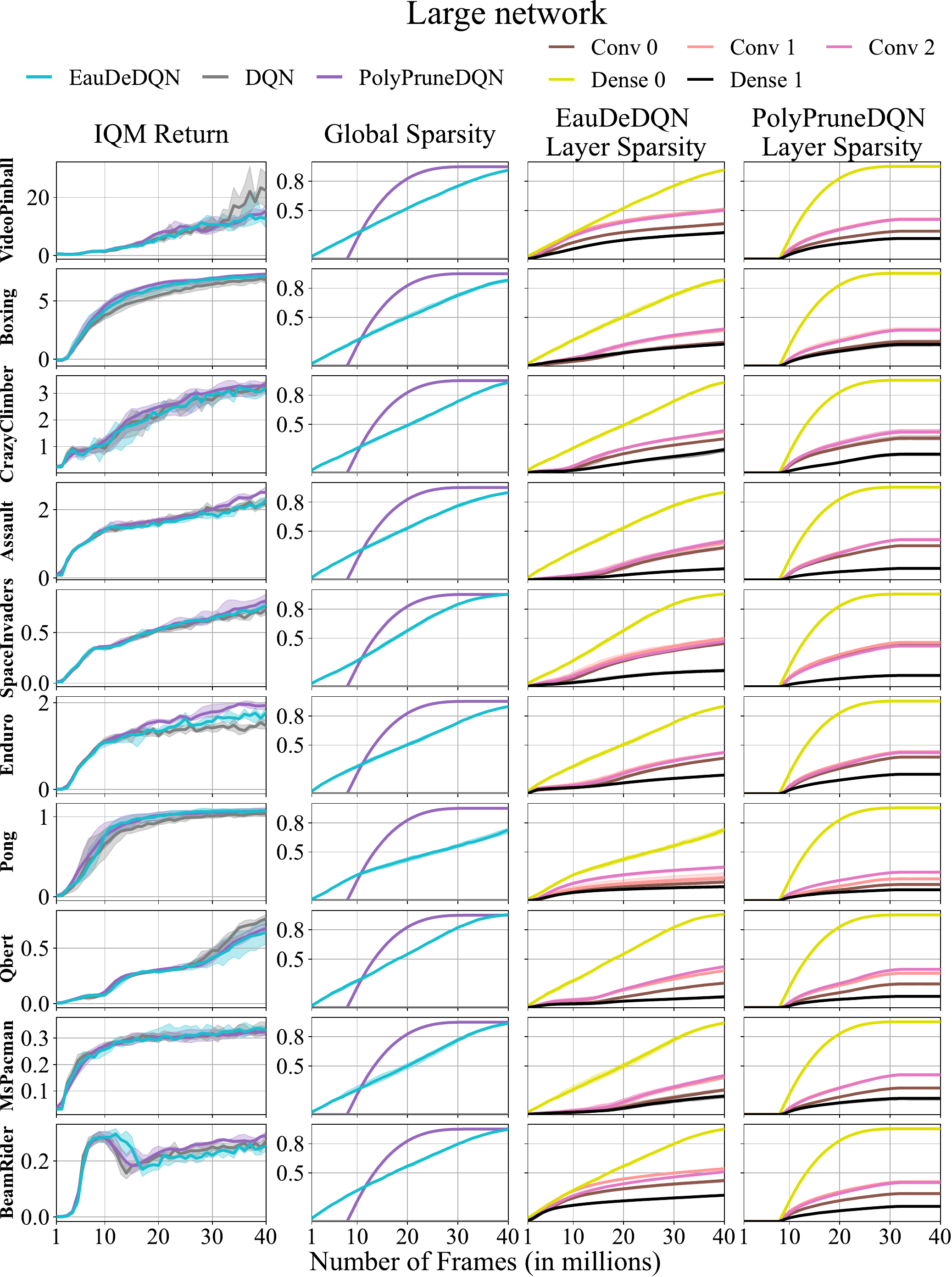}
    \caption{\textbf{Online Atari:} Per game metrics for the experiment on the large network. The aggregated performances are available in Figure~\ref{F:atari_online_performances} (top, right).}
\end{figure}

\begin{figure}[H]
    \centering
    \includegraphics[width=\textwidth]{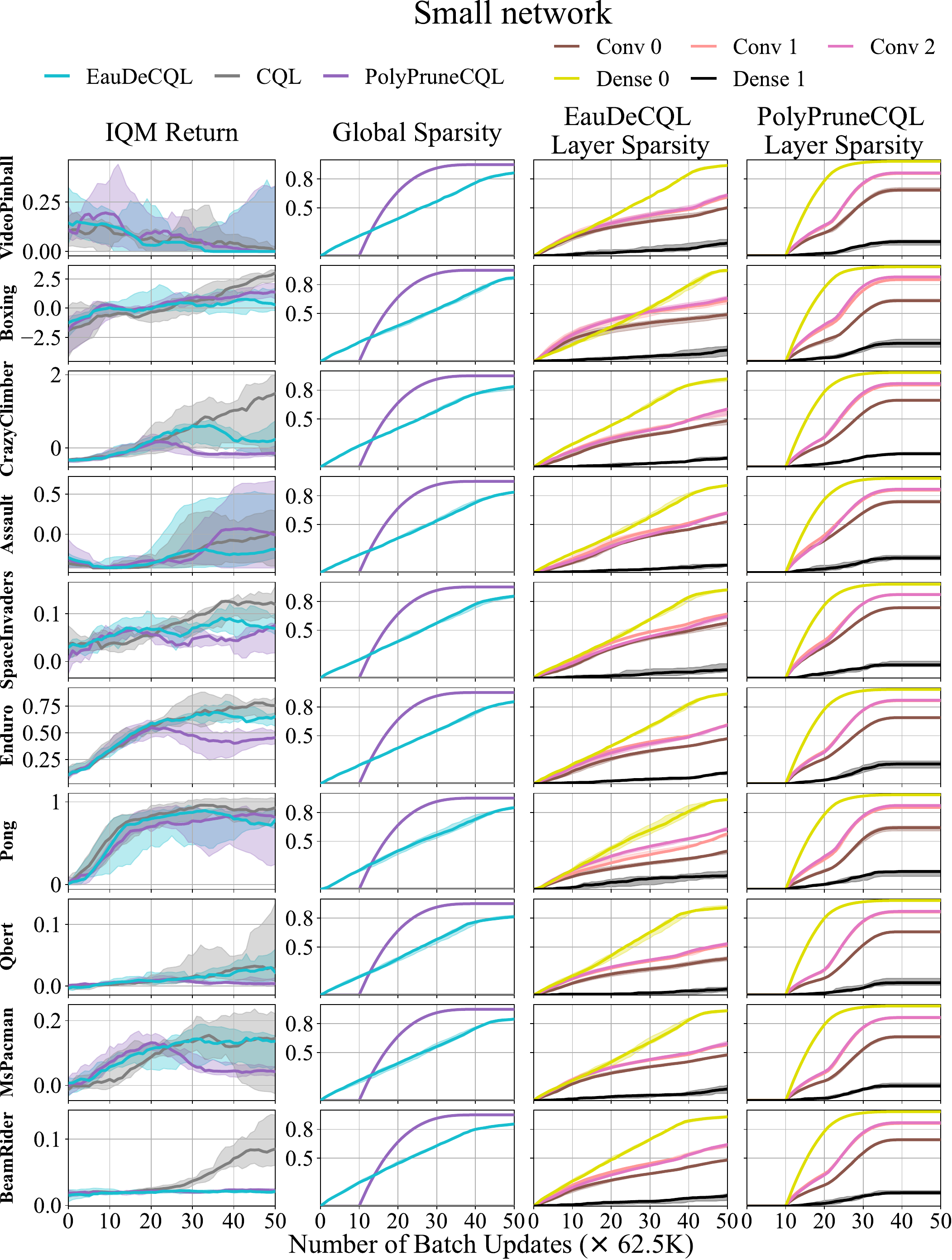}
    \caption{\textbf{Offline Atari:} Per game metrics for the experiment on the small network. The aggregated performances are available in Figure~\ref{F:atari_offline_performances} ($1^{\text{st}}$ plot to the left).}
\end{figure}
\begin{figure}[H]
    \centering
    \includegraphics[width=\textwidth]{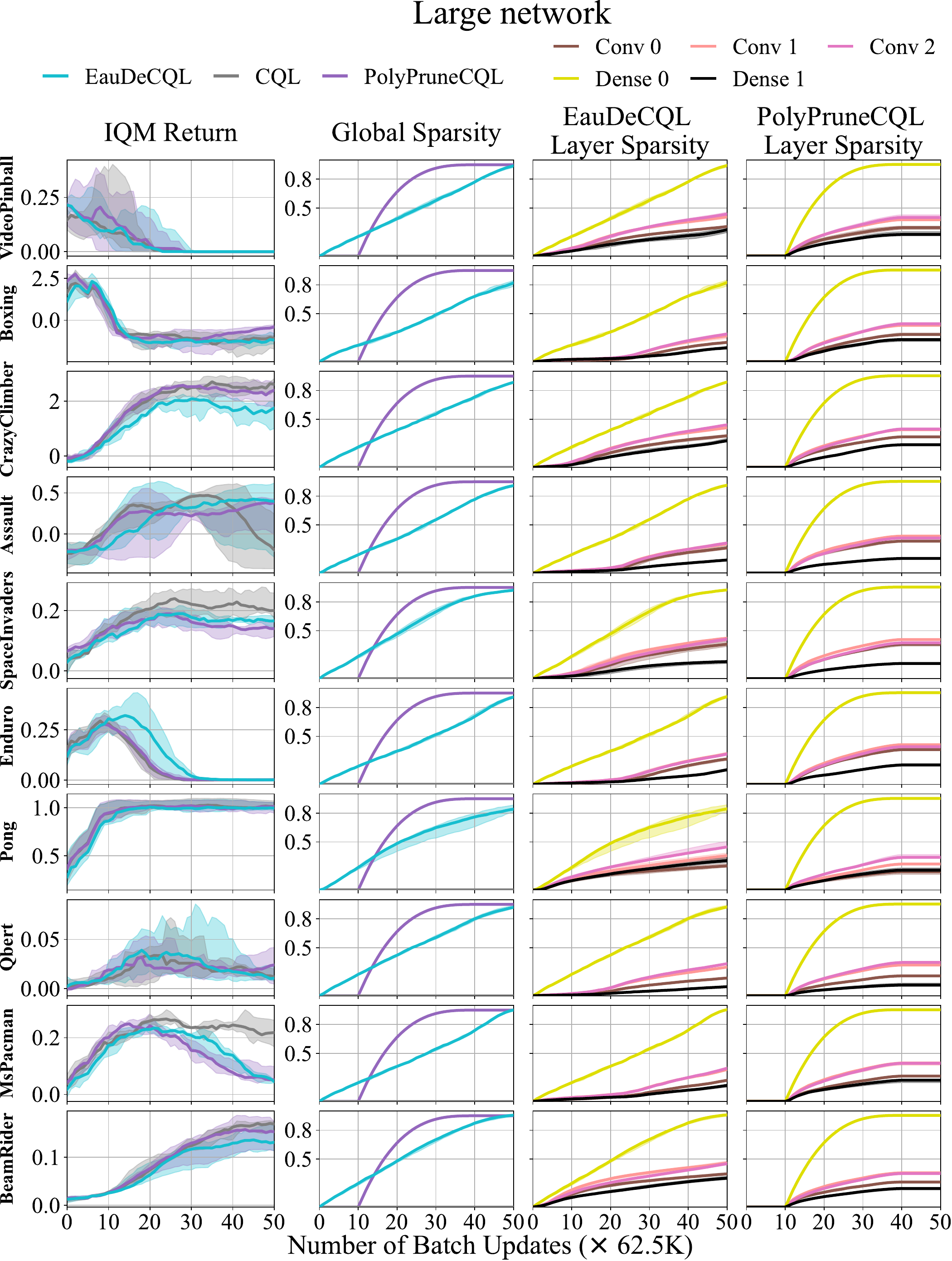}
    \caption{\textbf{Offline Atari:} Per game metrics for the experiment on the large network. The aggregated performances are available in Figure~\ref{F:atari_offline_performances} ($2^{\text{nd}}$ plot to the left).}
\end{figure}

\begin{figure}[H]
    \centering
    \includegraphics[width=\textwidth]{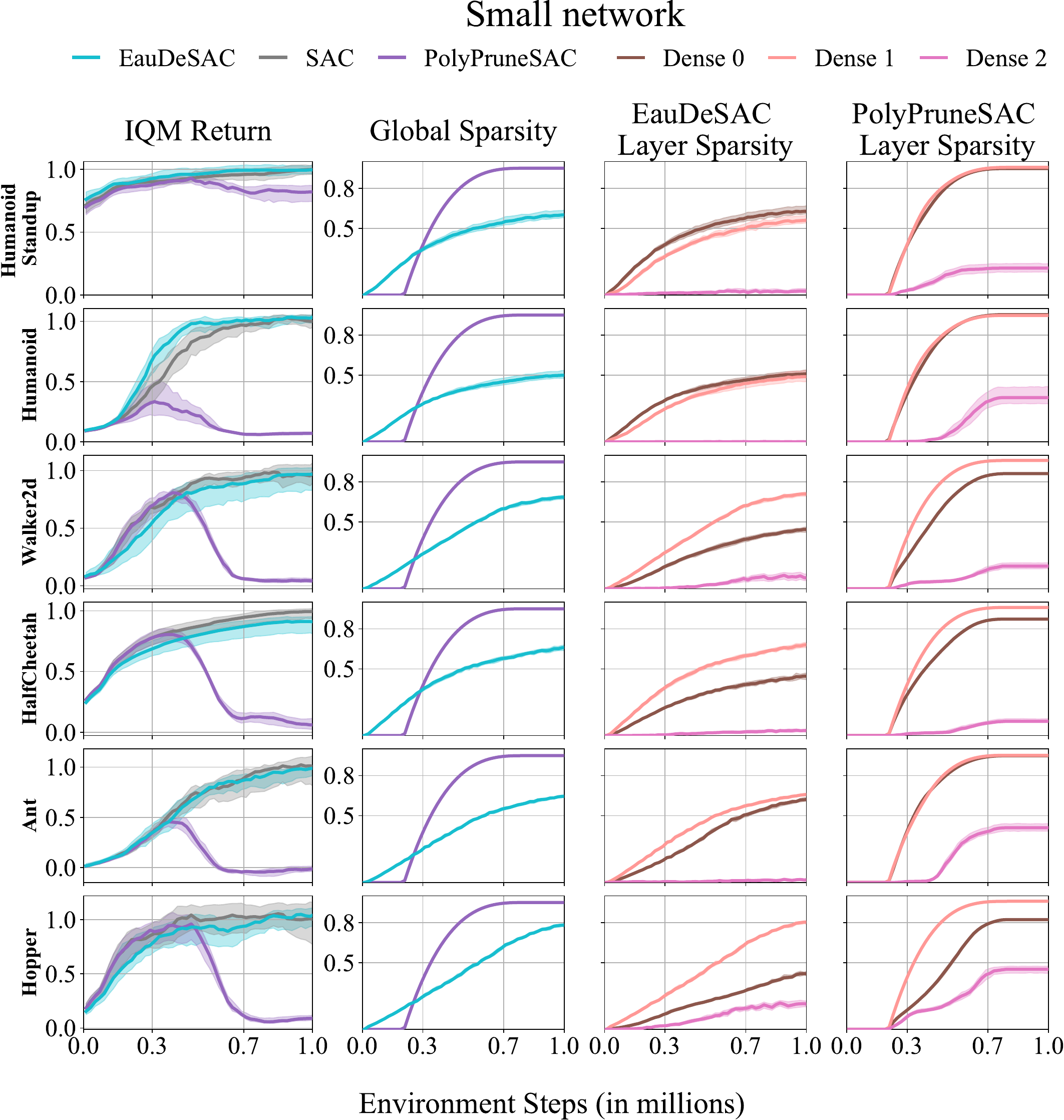}
    \caption{\textbf{Online MuJoCo:} Per game metrics for the experiment on the small network. The aggregated performances are available in Figure~\ref{F:mujoco_performances} (top, left).}
\end{figure}
\begin{figure}[H]
    \centering
    \includegraphics[width=\textwidth]{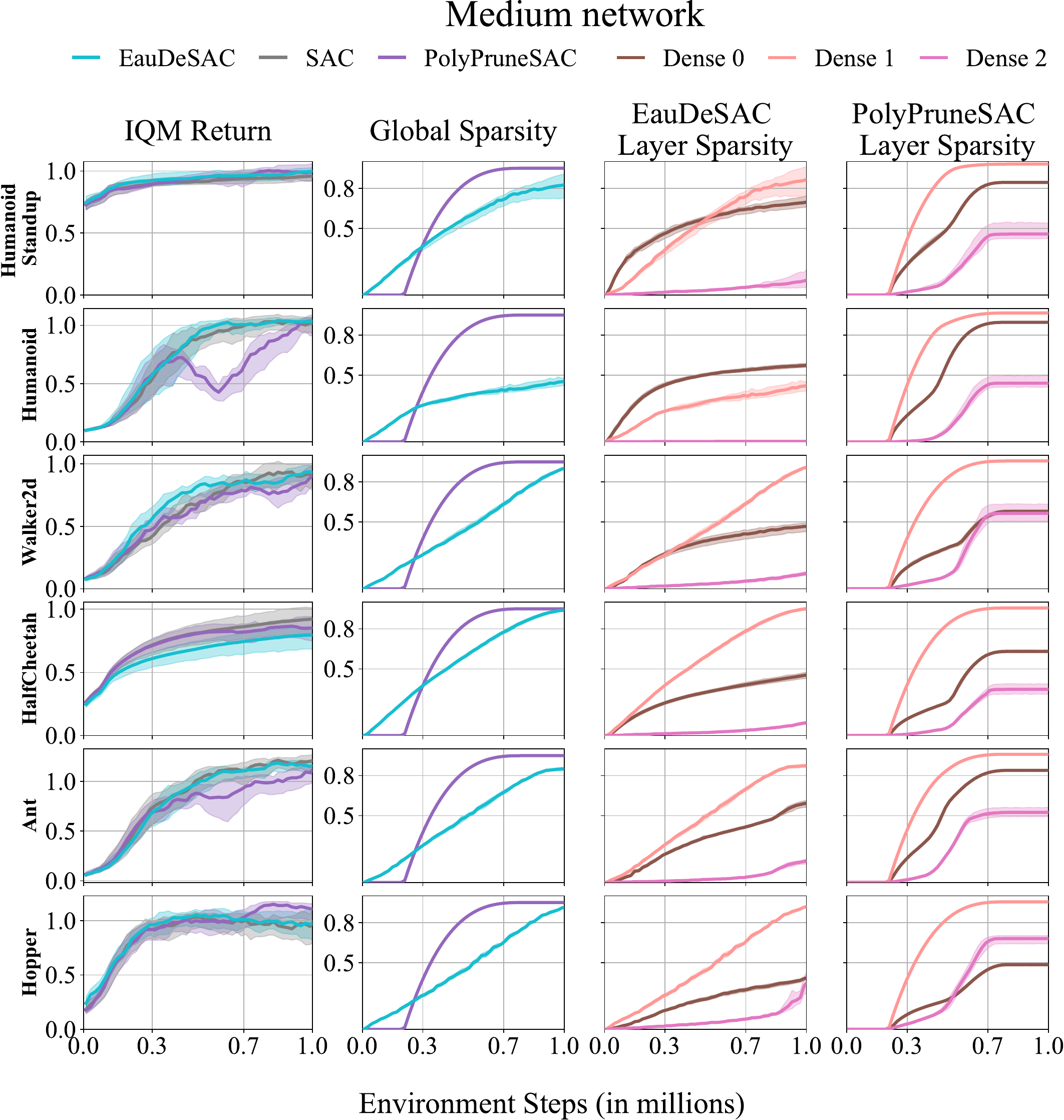}
    \caption{\textbf{Online MuJoCo:} Per game metrics for the experiment on the medium network. The aggregated performances are available in Figure~\ref{F:mujoco_performances} (top, middle).}
\end{figure}
\begin{figure}[H]
    \centering
    \includegraphics[width=\textwidth]{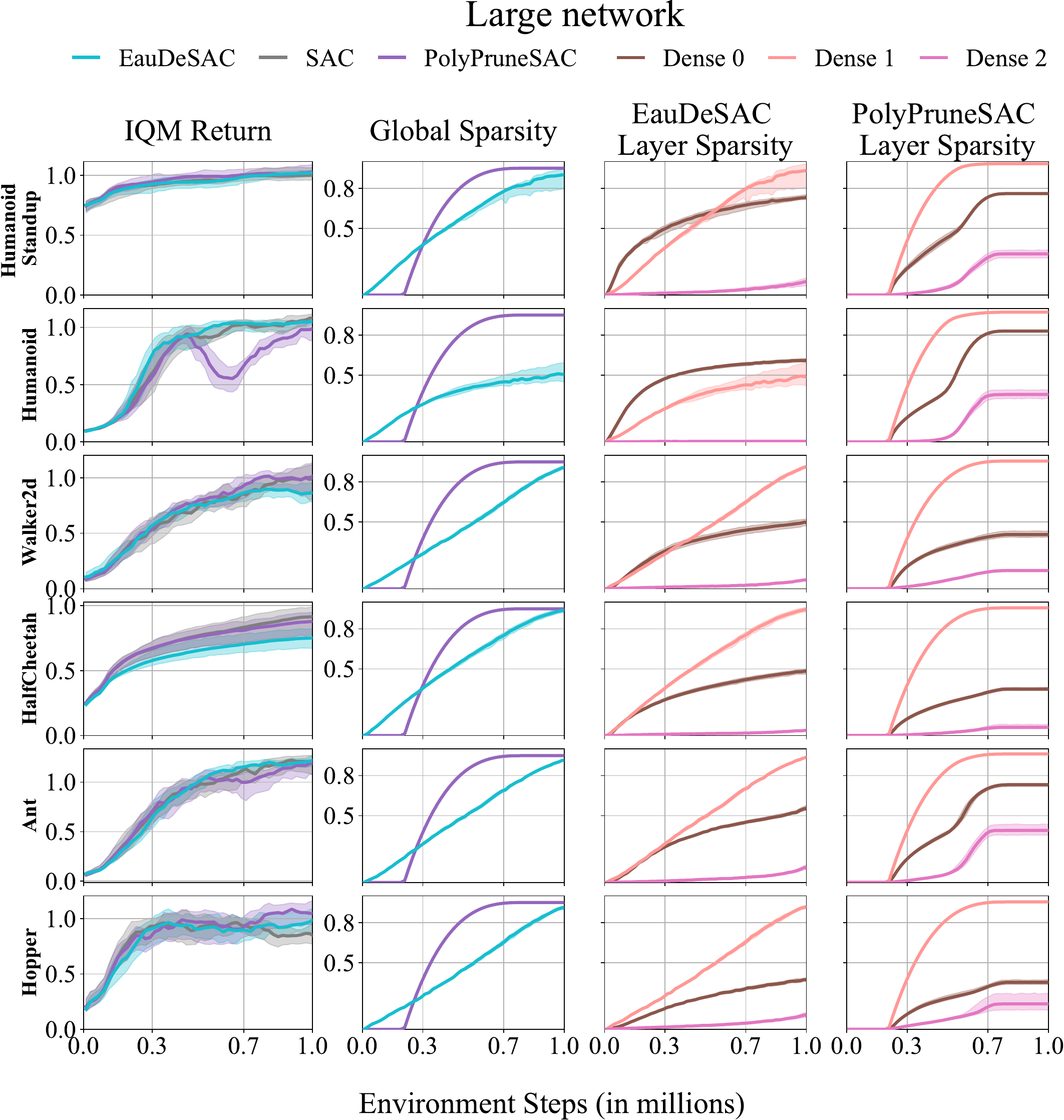}
    \caption{\textbf{Online MuJoCo:} Per game metrics for the experiment on the large network. The aggregated performances are available in Figure~\ref{F:mujoco_performances} (top, right).}
\end{figure}

\begin{table}
\begin{minipage}{0.49\textwidth}
    \centering
    \caption{Summary of the shared hyperparameters used for the Atari experiments. We note $\text{Conv}_{a,b}^d C$ a $2D$ convolutional layer with $C$ filters of size $a \times b$ and of stride $d$, and $\text{FC }E$ a fully connected layer with $E$ neurons.}\label{T:atari_parameters}
    \begin{tabular}{ l | r }
        \toprule
        \multicolumn{2}{ c }{Environment} \\ 
        \hline
        Discount factor $\gamma$ & $0.99$ \\
        \hline
        Horizon $H$ & $27\,000$ \\
        \hline
        Full action space & No \\
        \hline 
        Reward clipping & clip($-1, 1$) \\
        \midrule
        \multicolumn{2}{ c }{All experiments} \\ 
        \hline
        Batch size & $32$ \\
        \hline    
        \multirow{3}{*}{Torso architecture} & $\text{Conv}_{8,8}^4 32$ \\
        & $ - \text{Conv}_{4,4}^2 64$ \\
        & $ - \text{Conv}_{3,3}^1 64$ \\
        \hline
        \multirow{4}{*}{Head architecture} & $\text{FC }32$ (small) \\
        & $\text{FC }512$ (medium)  \\
        & $\text{FC }2048$ (large)  \\
        & $- \text{FC }n_{\mathcal{A}}$ \\
        \hline   
        Activations & ReLU \\
        \hline
        PolyPruneQN & $4\,000$ (online) \\
        pruning period & $1\,000$ (offline) \\
        \midrule
        \multicolumn{2}{ c }{Online experiments} \\ 
        \hline
        Number of training & \multirow{2}{*}{$250\,000$} \\
        steps per epochs & \\
        \hline
        Target update & \multirow{2}{*}{$8\,000$} \\
        period $T$ & \\
        \hline
        Type of the & \multirow{2}{*}{FIFO} \\
        replay buffer $\mathcal{D}$ & \\
        \hline 
        Initial number & \multirow{2}{*}{$20\,000$} \\
        of samples in $\mathcal{D}$ & \\
        \hline 
        Maximum number & \multirow{2}{*}{$1\,000\,000$} \\
        of samples in $\mathcal{D}$ & \\
        \hline
        Gradient step & \multirow{2}{*}{$4$} \\
        period $G$ & \\
        \hline    
        Starting $\epsilon$ & $1$ \\
        \hline    
        Ending $\epsilon$ & $0.01$ \\
        \hline
        $\epsilon$ linear decay & \multirow{2}{*}{$250\,000$} \\
        duration & \\
        \hline    
        Batch size & $32$ \\
        \hline    
        Learning rate & $6.25 \times 10^{-5}$ \\
        \hline    
        Adam $\epsilon$ & $1.5 \times 10^{-4}$ \\
        \midrule
        \multicolumn{2}{ c }{Offline experiments} \\ 
        \hline
        Number of training & \multirow{2}{*}{$62\,500$} \\
        steps per epochs & \\
        \hline
        Target update & \multirow{2}{*}{$2\,000$} \\
        period $T$ & \\
        \hline
        Dataset size & $2\,500\,000$ \\
        \hline     
        Learning rate & $5 \times 10^{-5}$ \\
        \hline    
        Adam $\epsilon$ & $3.125 \times 10^{-4}$ \\ 
        \bottomrule
    \end{tabular}
\end{minipage}
\hspace{0.03cm}
\begin{minipage}{0.49\textwidth}
    \centering
    \caption{Summary of all hyperparameters used for the MuJoCo experiments. We note $\text{FC }E$ a fully connected layer with $E$ neurons.} \label{T:mujoco_parameters}
    \begin{tabular}{ l | r }
        \toprule
        \multicolumn{2}{ c }{Environment} \\ 
        \hline
        Discount factor $\gamma$ & $0.99$ \\
        \hline
        Horizon $H$ & $1\,000$ \\
        \hline
        \multicolumn{2}{ c }{All algorithms} \\ 
        \hline
        Number of & \multirow{2}{*}{$1\,000\,000$} \\
        training steps & \\
        \hline
        Type of the & \multirow{2}{*}{FIFO} \\
        replay buffer $\mathcal{D}$ & \\
        \hline 
        Initial number & \multirow{2}{*}{$5\,000$} \\
        of samples in $\mathcal{D}$ & \\
        \hline 
        Maximum number & \multirow{2}{*}{$1\,000\,000$} \\
        of samples in $\mathcal{D}$ & \\
        \hline
        Update-To-Data & \multirow{2}{*}{$1$} \\
        UTD & \\
        \hline   
        Batch size & $256$ \\
        \hline    
        Learning rate & $10^{-3}$ \\
        \hline 
        Policy delay & $1$ \\
        \hline 
        \multirow{2}{*}{Actor architecture} & $\text{FC }256$ \\
        & $- \text{FC }256$ \\
        \hline 
        \multirow{6}{*}{Critic architecture} & $\text{FC }256$ \\
        & $- \text{FC }256$ (small) \\
        & $\text{FC }1280$ \\
        & $- \text{FC }1280$ (medium) \\
        & $\text{FC }2048$ \\
        & $- \text{FC }2048$ (large) \\
        \hline 
        Soft target update & \multirow{2}{*}{$5 \times 10^{-3}$} \\
        period $\tau$ & \\
        \hline
        Pruning period $P$ & $1\,000$ \\
        \bottomrule
    \end{tabular}
\end{minipage}
\end{table}

\end{document}